\documentclass[sn-mathphys]{sn-jnl}

\jyear{2021}%

\usepackage{booktabs}
\usepackage{diagbox}
\usepackage{color}
\usepackage{amsmath,amssymb}
\usepackage{graphicx}

\theoremstyle{thmstyleone}%
\theoremstyle{thmstyletwo}%

\theoremstyle{thmstylethree}%
\newtheorem{definition}{Definition}%

\begin{document}

\title[Article Title]{An Empirical Study of Pre-trained Language Models in Simple Knowledge Graph Question Answering}


\author[1]{\fnm{Nan} \sur{Hu}}\email{nanhu@seu.edu.cn}

\author[1]{\fnm{Yike} \sur{Wu}}\email{wuyike@seu.edu.cn}

\author*[1]{\fnm{Guilin} \sur{Qi}}\email{gqi@seu.edu.cn}

\author[1]{\fnm{Dehai} \sur{Min}}\email{zhishanq@seu.edu.cn}

\author[2]{\fnm{Jiaoyan} \sur{Chen}}\email{jiaoyan.chen@manchester.ac.uk}

\author[3]{\fnm{Jeff Z.} \sur{Pan}}\email{j.z.pan@ed.ac.uk}

\author[1]{\fnm{Zafar} \sur{Ali}}\email{zafarali@seu.edu.cn}

\affil*[1]{\orgdiv{School of Computer Science and Engineering}, \orgname{Southeast University}, \orgaddress{\street{2 Dongda Rd}, \city{Nanjing}, \postcode{211189}, \state{Jiangsu}, \country{China}}}

\affil[2]{\orgdiv{Department of Computer Science}, \orgname{The University of Manchester}, \orgaddress{\street{Oxford Rd}, \city{Manchester}, \postcode{M13 9PL},\country{UK}}}

\affil[3]{\orgdiv{School of Informatics}, \orgname{The University of Edinburgh}, \orgaddress{\street{10 Crichton St}, \city{Edinburgh}, \postcode{EH8 9AB}, \country{UK}}}


\abstract{
Large-scale pre-trained language models (PLMs) such as BERT have recently achieved great success and become a milestone in natural language processing (NLP). It is now the consensus of the NLP community to adopt PLMs as the backbone for downstream tasks. In recent works on knowledge graph question answering (KGQA), BERT or its variants have become necessary in their KGQA models.
However, there is still a lack of comprehensive research and comparison of the performance of different PLMs in KGQA. To this end, we summarize two basic KGQA frameworks based on PLMs without additional neural network modules to compare the performance of nine PLMs in terms of accuracy and efficiency. In addition, we present three benchmarks for larger-scale KGs based on the popular SimpleQuestions benchmark to investigate the scalability of PLMs. We carefully analyze the results of all PLMs-based KGQA basic frameworks on these benchmarks and two other popular datasets, WebQuestionSP and FreebaseQA, and find that knowledge distillation techniques and knowledge enhancement methods in PLMs are promising for KGQA. Furthermore, we test ChatGPT\footnote{https://chat.openai.com/}, which has drawn a great deal of attention in the NLP community, demonstrating its impressive capabilities and limitations in zero-shot KGQA.} We have released the code and benchmarks to promote the use of PLMs on KGQA.\footnote{https://github.com/aannonymouuss/PLMs-in-Practical-KBQA}




\maketitle

\section{Introduction}\label{sec1}

As a milestone work in the field of NLP, substantial work has shown that the pre-trained language models (PLMs) with self-supervised learning on large-scale corpora have learned the rich semantic knowledge that facilitates a variety of NLP downstream tasks \cite{Pre-trainedModels}.
Manning \cite{ref_article1} explains that PLMs learn the meaning of words because meaning can be considered as a connection of linguistic forms, and PLMs have seen many connections of words to understand the meaning of words. For example, PLMs understand the word ``Washington, D.C." by these two sentences ``Washington, D.C. is the capital city and federal district of the United States'' and ``Washington, D.C. is located on the east bank of the Potomac River". As a result, PLMs significantly improve the performance of most NLP tasks through such learning.

Today, it has become a consensus to use PLMs, through fine-tuning or prompting, as the backbone of downstream tasks.
With the proposed initial PLMs BERT \cite{BERT} and GPT \cite{GPT}, the PLMs community began to flourish. Subsequently, various PLMs are proposed to address different needs and tasks.
For example, some works propose knowledge distillation models such as DistilBERT \cite{DistilBERT} and TinyBERT \cite{TinyBERT} to reduce the number of parameters in PLMs and to increase the speed of training and inference. Some works present knowledge-enhanced models such as ERNIE \cite{Zhang2019} and KEPLER \cite{KEPLER2021} to address knowledge-driven downstream tasks. However, there is a lack of a comprehensive study of the application of these PLMs to an important sub-task of NLP, namely knowledge graph question answering (KGQA).

KGQA aims to find answers to natural language questions from the knowledge graph (KG) which is typically store structured knowledge in the form of triples, denoted as (\textit{subject}, \textit{relation}, \textit{object}). 
The study of various PLMs applied to KGQA is valuable for the following reasons.

\begin{figure}
\begin{center}{
\includegraphics[width=0.95\textwidth]{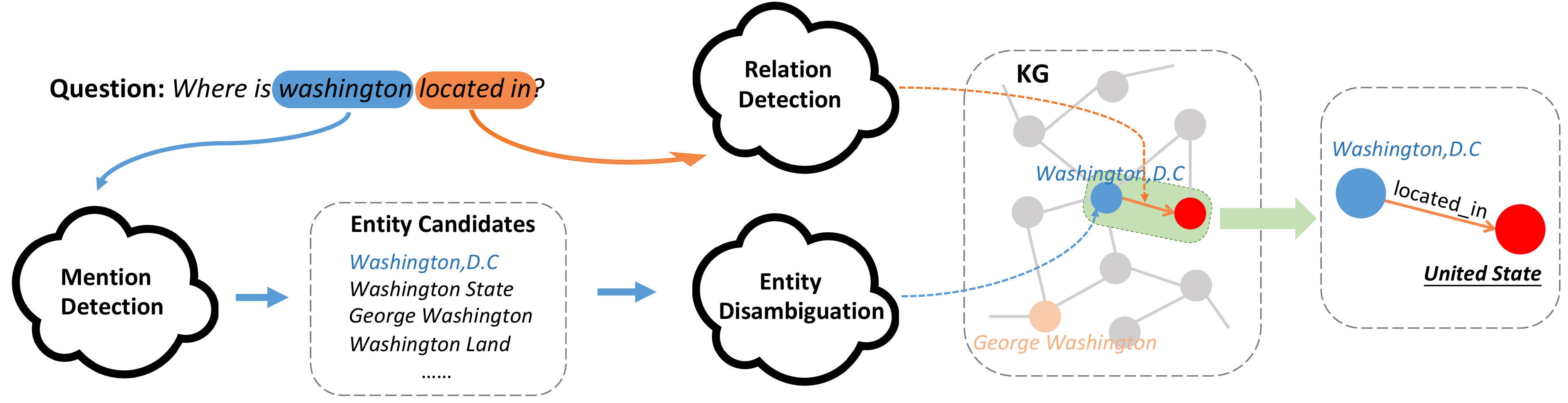}
\caption{The general structure of KGQA.} \label{figure-1}
}
\end{center}
\end{figure}

\begin{enumerate}[1.]
\item The general domain KGQA is difficult to use in practice because of the efficiency issue.
The huge size of the general domain KG\footnote{Freebase \cite{Bollacker2008} contains over 3 billion triples across over 100 domains, while Google Knowledge Graph \cite{Danny2020} has amassed over 500 billion triples.}
leads to a large training and inference time for the KGQA system. 
Some KGQA works limit the search range in KG to reduce complexity \cite{Lan2020,Gu2021,Ye2022}, yet the training time for a well-performing KGQA model \cite{Lan2020} still exceeds 2 weeks. Without this search limitation, the model \cite{Lan2020} would even take a few months to train as noted in \cite{Gu2021}. Several works have attempted to reduce complexity by optimizing KGQA approaches \cite{Gu2022,Chen2021,Qin2021}, but at the expense of performance. 
In the recent KGQA system, PLMs have become a regular component of the system due to their obvious gains in performance.
However, the application of powerful but large PLMs further increases the difficulty of using KGQA systems in practice. \textbf{Therefore, it is necessary to explore the trade-off between performance and efficiency of PLMs on KGQA}.

\item KGQA is a knowledge-intensive task and involves several common NLP subtasks\footnote{These tasks are similar to named entity recognition, entity linking and relation extraction.}
such as mention detection, entity disambiguation and relation detection as shown in Fig.\ref{figure-1}. There are substantial works to tackle these subtasks using PLMs as the cornerstone and with success \cite{Li2020,Wu2020,Chen2020,Oliya2021,Wang2021}. Moreover, several works utilize structured knowledge to enhance PLMs for more than just self-supervised training on the large-scale corpus, with significant improvements in mention detection \cite{Yamada2020} and relation detection \cite{Zhang2022,Zhang2019,KEPLER2021,Peters2019} tasks. 
Nevertheless , there is lack of work on making a comprehensive comparison on each subtasks from the perspective of PLMs. \textbf{Therefore, exploring the use of PLMs for KGQA is also instructive for these subtasks}.

\end{enumerate}

This work aims to evaluate the overall performance of various PLMs on KGQA comprehensively.
We not only examine the \textbf{accuracy} and \textbf{efficiency} metrics of KGQA systems based on different PLMs, but also study their \textbf{scalability}\footnote{Scalability is the measure of a system’s ability to increase or decrease in performance and cost in response to changes in system processing demands. In our work, we explore the variation in accuracy performance and time cost with increasing KG size.}.
Specifically, we designed four KGs of increasing size to explore the variation of these KGQA systems. \textbf{Three} classes of \textbf{nine} PLMs are used for evaluation, including the common PLMs \textsc{Bert} \cite{BERT}, \textsc{Roberta} \cite{RoBERTa}, \textsc{XLnet} \cite{XLnet} and \textsc{Gpt2} \cite{GPT}, the lightweight PLMs \textsc{ALbert} \cite{ALBERT2021}, \textsc{DistilBert} \cite{DistilBERT} and \textsc{DistilRoberta} \cite{DistilBERT}, and the knowledge-enhanced PLMs \textsc{Luke} \cite{Yamada2020} and \textsc{Kepler} \cite{KEPLER2021}. 
As common models serve as the backbone models for lightweight PLMs and knowledge-enhanced PLMs, we follow \cite{Pre-trainedModels} to further classify them according to pre-trained task categories, namely, masked language modeling (i.e., \textsc{Bert} and \textsc{Roberta}), language modeling (i.e., \textsc{Gpt2}), and permuted language modeling (i.e., \textsc{XLnet}).
The investigation focused on the simple but common Simple KGQA task, which can be answered by a triple in KG. Moreover, we summarize two basic KGQA frameworks from previous works for the experiment. These two frameworks are vanilla without additional neural network modules except for PLMs and simple linear layers. It allows us to focus on comparing PLMs instead of various KGQA approaches with complex neural network modules. We also conducte experiments to compare the performance of these PLMs under fine-tuning with ChatGPT under zero-shot settings on the KGQA task.

In summary, our main contributions are as follows.

\begin{enumerate}[1.]
\item To the best of our knowledge, this is the first attempt to comprehensively study the overall performance of various PLMs in KGQA tasks. 
For this purpose, we summarize two basic KGQA frameworks from popular simple KGQA approaches to exclude the interference of complex neural network modules. We have implemented 18 KGQA systems based on these two basic KGQA frameworks using a total of nine PLMs. Further, we propose three KGQA benchmarks based on the popular SimpleQuestions benchmark. These four benchmarks have a linearly increasing KG scale.

\item We conduct comprehensive experiments to evaluate the overall results of all implemented KGQA systems on all benchmarks. We make detailed analyses regarding overall accuracy, efficiency and scalability from the perspective of different PLMs and KGQA frameworks. In addition, we further analyze the overall performance of the sub-modules of the KGQA systems to investigate the impact of the different PLMs and KGQA frameworks on these subtasks. We also compare the performance of these PLMs under fine-tuning with ChatGPT under zero-shot settings on three KGQA datasets.

\item We find that knowledge-distilled lightweight PLMs and knowledge-enhanced PLMs are promising for use in KGQA. This leads us to delve into this direction in the future to explore practical KGQA systems. Besides, we observe that ChatGPT has an excellent performance in KGQA tasks, while there are still some limitations. Section \ref{Sec6} summarizes all the important findings.
Our KGQA frameworks based on PLMs provide new strong baselines of simple KGQA. We have released code and benchmarks as publicly accessible resources to help the future development of the KGQA community.

\end{enumerate}

The rest of the article is structured as follows. In Section \ref{RelatedWorks}, we introduce related works on simple KGQA and PLM Applications On KGQA. In Section \ref{Sec2}, we present the preliminary knowledge of this work. In Section \ref{Sec3}, we summarise the existing simple KGQA methods and describe the two summarised KGQA basic frameworks in detail. In Section \ref{Sec4}, we introduce four benchmarks and evaluation metrics. In Section \ref{Sec5}, we describe the results and analysis of the experiment. Finally, Section \ref{Sec6} concludes this work and introduces future works. 

\section{Related Works}\label{RelatedWorks}

\subsection{Simple Knowledge Graph Question Answering}

Knowledge graph question answering (KGQA) aims to find answers to natural language questions from the knowledge graph (KG). Simple KGQA means that a natural language question can be answered by a triplet in KG. The two mainstream branches of the current KGQA methods are information retrieval (IR) and semantic parsing (SP) \cite{AReview,ComplexKnowledge,KnowledgeBase}. The former attempts to retrieve answers directly from a subgraph centred on the topic entity and then models answer features for ranking. The latter tries to train a semantic parser to transform the question into intermediate logical forms and then execute it against KG. In simple KGQA, the IR method employs various neural networks to score the similarity between the question and each candidate fact in the subgraph and then find the best match. It follows the process of \textit{retrieving} question-specific subgraph and then \textit{ranking} the facts in it. Bordes et al. \cite{ref_article25} used a memory network to encode questions and facts into the same vector space and score their similarity. Dai et al. \cite{ref_article15} proposed a two-step conditional probability estimation problem and adopted a BiGRU network as an encoder. Yu et al. \cite{Yu2017} designed two independent hierarchical residual BiLSTMs to represent questions and relations with different granularities. Yin et al. \cite{ref_article11} used two independent models, a character-level CNN and a word-level CNN with attentive max-pooling. Lukovnikov et al. \cite{ref_article9} proposed an end-to-end word/character-level encoding network for ranking subject-relation pairs and retrieving relevant facts. In simple KGQA, the SP method is simplified to a \textit{classification} model because only a relation or a predicate needs to be generated. Ture and Jojic \cite{ref_article6} employed a two-layer BiGRU model as a classifier. Petrochuk and Zettlemoyer \cite{ref_article5} used a BiLSTM to classify relations and achieve state-of-the-art performance. Mohammed et al. \cite{ref_article2} only adopted simple neural networks (i.e. LSTMs and GRUs) or non-neural network models (i.e. CRFs).

In Section 4, we name IR method and SP method as the \textit{retrieval and ranking-based} method and the \textit{classification-based} method to show the differences more clearly.

\subsection{PLM-based Methods for KGQA}

Pretrained language models have been widely served for various downstream tasks, including KGQA, due to the powerful representation capabilities learned from large-scale text corpora. For the IR method, PLMs provide a unified way to model unstructured text and structured KG information in a unified semantic space, which facilitates question-specific subgraph reasoning. Zhang et al. \cite{SubgraphRetrieval} trained a PLM-based path retriever to retrieve hop-by-hop question-related relations. At each step, the retriever ranked the top-k relations based on the question and the relations selected in the previous step. Hu et al. \cite{ImprovingCore} introduced PLM to help align questions and paths in a step-wise reasoning manner from explicit text semantic matching and implicit KG structure matching. Luo et al. \cite{ref_article14} proposed a BERT-based model to preserve the original question-fact interaction information and reduce the semantic gap.
For the SP method, PLMs significantly improve the understanding of questions, especially complex ones. Lukovnikov et al. \cite{ref_article3} made the first attempt to use PLMs as classifiers to predict relations, with a significant performance improvement over shallow neural networks. In addition, Lukovnikov et al. demonstrated the greater advantage of PLMs on limited training data. Some works \cite{Case-based,RNG-KBQA} used PLMs to directly generate executable programs based on a given question and other relevant KG information. Substantial improvement in model performance demonstrates the effectiveness of such usages of PLMs. However, few KGQA works have taken into account the efficiency of PLMs. This is crucial for KGQA, which is inherently difficult to apply in practice.

\section{Preliminaries}\label{Sec2}
In this section, we introduce the definition of simple KGQA task (Section \ref{Sec2.1}) and large-scale pre-trained language models (PLMs) (Section \ref{Sec2.2}).

\subsection{Task Definition}\label{Sec2.1}
This work focuses on evaluating PLMs on simple knowledge graph question answering, where the natural language question can be answered by a triple in KG. Simple questions are frequently queried in search engines and question-answering robots. The 100 most frequently asked questions on Google search engines in 2021 are simple questions\footnote{Available online: https://keywordtool.io/blog/most-asked-questions/ (accessed on 12 April 2022).}, and most of them can be answered by the KGQA system.

For ease of understanding, we define some notations used in this paper. 
Formally, a knowledge graph (KG) is typically a collection of subject-relation-object triples, denoted by $\mathbb{G}=\left\{(s, r, o)\vert s,o \in \mathbb{E}, r \in \mathbb{R} \right\}$, where $(s, r, o)$ denotes that relation $r$ exists between subject $s$ and object $o$, $\mathbb{E}$ and $\mathbb{R}$ denote the entity set and relation set, respectively.
Given the available KG $\mathbb{G}$, KGQA aims to answer natural language questions $Q = \left\{ w_1, w_2, ..., w_n \right\}$ in the format of a sequence of words with the answers $\mathcal{A}_q \subset \mathbb{E}$. 
For simple KGQA task, the answers directly connect to the topic entity and a KGQA system is trained using a dataset $D = \left\{ Q, \left \langle s, r\right \rangle \right\}$, where $\left \langle s, r\right \rangle$ refers to a subject-relation pair.

In inference stage, given a natural language question $Q$ ``\textit{The film Forrest Gump is directed by who?}" as shown in Fig.\ref{figure-1} , the KGQA system can answer this question by the answer \textit{Robert Zemeckis} which is retrieved by the subject-relation pair $\left \langle \textit{Forrest Gump, directed\_by} \right \rangle$ in the KG $\mathbb{G}$.

\subsection{Large-scale Pretrained Language Models}\label{Sec2.2}
Neural network-based language models represent everything through vectors of real numbers. They can learn better representations on a large corpus by back-propagating from the language model prediction task to the representation of words. Early work on language models trained shallow networks to capture the semantic meaning of words, such as Word2Vec \cite{Word2Vec} and GloVe \cite{GloVe}. However, they suffer from the drawback of not being able to represent polysemantic words in different contexts.

Since the introduction of Transformer \cite{Transformer}, it has become feasible to train deep neural models for NLP tasks. Transformer is a more complex model than the simple neural networks previously explored by humans for word sequences. One of its main ideas is the attention mechanism, by which the representation of one location is computed as a weighted combination of representations from other locations. With Transformer as the architecture, various PLMs trained on large-scale corpora such as BERT \cite{BERT} and RoBERTa \cite{RoBERTa} have been proposed with the goal of language model learning. Large-scale PLMs with hundreds of millions of parameters can learn polysemantic words as well as factual knowledge from contextual semantics. Furthermore, numerous works have proposed the use of structured knowledge to enhance PLMs, such as KEPLER \cite{KEPLER2021}, and others have used distillation techniques to reduce the number of PLM parameters, such as DistilBERT \cite{DistilBERT}. Most of them are based on improvements of the primary PLMs, i. e., BERT and RoBERTa. One such large-scale pre-trained language model can be deployed for many specific NLP tasks, requiring only a small number of further instructions. A standard approach is to fine-tune the model with a small amount of additional supervised learning. By fine-tuning large-scale PLMs, the rich linguistic knowledge of PLMs shows great performance on downstream NLP tasks.
Recently, ChatGPT PLM, released by OpenAI, has gained huge attention from the NLP community and many other fields. ChatGPT is fine-tuned from the GPT-3.5 series models through reinforcement learning from human feedback \cite{DeepReinforcement}. Several works \cite{IsChatGPT,AMultitask,HowClose} have shown that ChatGPT demonstrates powerful capabilities on a lot of NLP tasks, but testing in knowledge-intensive downstream tasks is lacking.
This work aims to explore the practicability of various PLMs on knowledge-intensive downstream tasks, i.e., knowledge graph question answering, to help researchers select the appropriate PLMs according to their needs. Unfortunately, ChatGPT currently only supports limited access ways and times, limiting our testing. We will leave more work about ChatGPT for the future.

\section{Two Basic KGQA Frameworks}\label{Sec3}

\subsection{Summary of the Framework}\label{FrameworkSummary}

To analyze the practicality of PLMs applied to KGQA. We summarise several simple KGQA approaches and propose two basic KGQA frameworks for evaluation\footnote{There is an existing KGQA approach based on KG embedding, which introduces knowledge representation learning, is proposed by \cite{KGEmbedding2019} and is not included in our frameworks. This work focuses on comparing various PLMs, so the discussion of the effect of different KG embedding methods is reserved for future work.}, a \textsc{cl}assification-based KGQA framework (\textsc{KGQAcl}) and a \textsc{r}etrieval and \textsc{r}anking-based KGQA framework (\textsc{KGQArr}). Previous works 
\cite{ref_article2,ref_article3,ref_article4,ref_article5,ref_article6}
belonging \textsc{KGQAcl} designed various deep neural networks to encode the question and then map the question vector to the KG relational dictionary. Previous works
\cite{Yu2017,ref_article8,ref_article9,ref_article10,ref_article11,ref_article12,ref_article13,ref_article14,ref_article15,ref_article16}
belonging \textsc{KGQArr} first retrieved adjacent relations (one-hop) of linked entities and then designed new network architecture or introduced contextual information to rank these relations. Some works also propose approaches such as utilize relation detection models to reorder entities \cite{Yu2017} or adopt a joint training strategy 
\cite{ref_article10,ref_article13}
to improve performance but make the KGQA framework more complex. Our frameworks do not consider these approaches as improving the performance of KGQA is not the purpose of this work.

Both basic frameworks consist of four modules, including (1) \textbf{Mention Detection}, (2) \textbf{Entity 
Disambiguation}, (3) \textbf{Relation Detection} and (4) \textbf{Answer Query}. The main difference between these two basic frameworks is the relation detection module. For \textsc{KGQArr}, this module is intended to to rank candidate relations (i.e. information retrieval). For \textsc{KGQAcl}, this module aims to map question intent to KG relations (i.e. semantic parsing).
Mention detection and entity disambiguation are also regarded as two steps of the entity linking task. 
Existing studies on KGQA typically treat entity linking as an individual task to be handled in advance \cite{KnowledgeBase}.
While \textsc{KGQAcl} usually treats entity linking and relation detection as separate modules, \textsc{KGQArr} considers the whole process as a pipeline, with relation detection coming after entity linking.

\subsubsection{Mention Detection}
Given a natural language question, the model will first find the mention representing the entity's name in that question. Previous works usually treated mention detection as a named entity recognition task and employed various models such as RNN, CNN and their variants
\cite{ref_article2,ref_article5,ref_article6,ref_article12}
or BERT \cite{ref_article14} to solve it. Other work regards it as span detection task \cite{ref_article3} or adopts CNN-LSTM as an encoder-decoder to generate entities directly \cite{ref_article4}.

\subsubsection{Entity Disambiguation}
The detected mentions will be used to collect candidate entities, and these candidates will then be ranked. Several works employ n-gram heuristics approaches 
\cite{ref_article2,ref_article3,ref_article6}
to collect entities efficiently, and then different methods such as  character similarity 
\cite{ref_article2,ref_article3,ref_article4},
TF-IDF scores \cite{ref_article6} are employed for entity disambiguation. \cite{ref_article5} disambiguate candidate entities by such a simple method as the score of connected relations. Our frameworks have adopted this simple method.

\subsubsection{Relation Detection} 
This module aims to obtain the correct relation in KG corresponding to the question. We summarise the two mainstream approaches, viz., \textsc{KGQAcl} and \textsc{KGQArr} in this work. The former is based on the idea of classification and maps the questions directly into the KG relational dictionary as being independent of the previous modules. Previous works use various models like RNN \cite{ref_article6}, LSTM 
\cite{ref_article2,ref_article4,ref_article5}
and BERT \cite{ref_article3} to encode sequences of questions, which are then classified into KG relation categories. 
The latter can be regarded as a similarity matching task, which will use linked entities to retrieve a set of candidate relations and then select the one with the highest similarity to the question. Various models 
\cite{Yu2017,ref_article8,ref_article25,ref_article14,ref_article15}, attention mechanism \cite{ref_article11,ref_article12} and external features such as context \cite{ref_article16} and type \cite{ref_article10} are designed to enhance the performance.

\subsubsection{Answer Query} Candidate entities and candidate relations with scores will be combined into pairs to query in KG to get the answers. The combination with the highest total score of the weighted sum of entities and relations is considered the correct pair 
\cite{ref_article2,ref_article3,ref_article4,ref_article5,ref_article6}.
Some works 
\cite{ref_article9,ref_article10,ref_article12,ref_article13}
jointly train entity disambiguation and relation detection to select the pair with the highest model score. However, this approach cannot be implemented in our PLMs-based frameworks due to the limitations of the GPU\footnote{Our basic frameworks are trained using an NVIDIA GeForce RTX 2080 TI}. So we do not consider this approach, which is difficult to ground
because of its excessive hardware requirements.

Note that the two basic KGQA frameworks we summarised are vanilla and contain only PLMs and simple linear layers. This makes sense as it allows us to focus on comparing PLMs. Except for answer query, other modules are implemented based on PLMs. In addition, the modules are identical in both frameworks, excluding the relation detection.
Next, we will detail these two PLMs-based KGQA frameworks.

\subsection{The Basic Classification-based KGQA Framework} 

\begin{figure}
\begin{center}{
\includegraphics[width=0.85\textwidth]{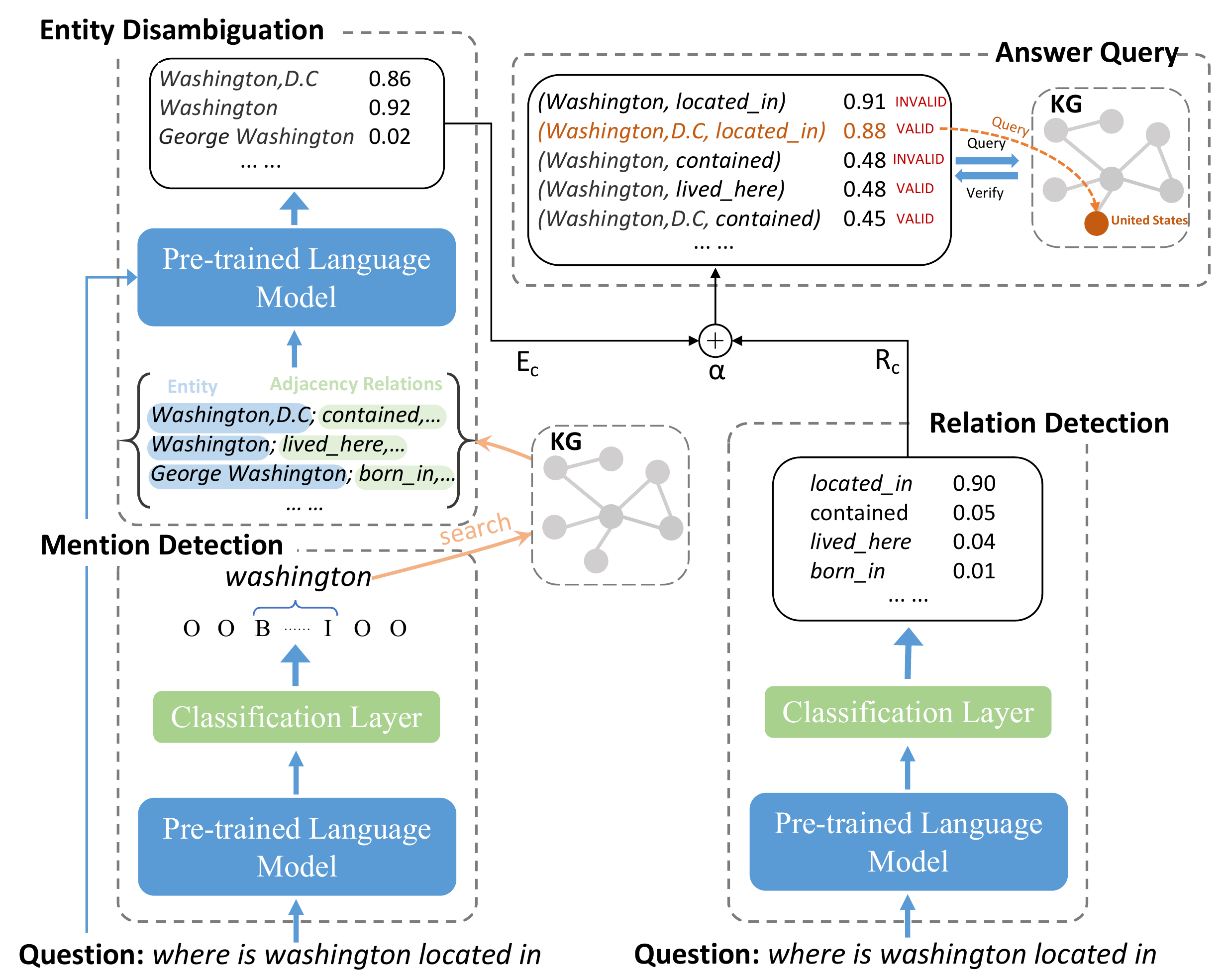}
\caption{The basic classification-based KGQA framework.} \label{CLFramework}
}
\end{center}
\end{figure}

The basic \textbf{CL}assification-based \textbf{KGQA} framework (\textsc{KGQAcl}) is shown in Fig.\ref{CLFramework}. It consists of four modules described in Section \ref{FrameworkSummary}, namely Mention Detection, Entity Disambiguation, Relation Detection and Answer Query. 

\subsubsection{Mention Detection} Given a question $Q$, the goal of mention detection is to identify the subject mention $m$. For instance, the subject mention of the question in Fig.\ref{CLFramework}  is ``$washington$".
We treat this task as a common PLMs-based named entity recognition task. The sequence of question is encoded by PLMs and will then be fed into a linear classification layer.
It will assign a label for each word in the question sequence, $B$ for the beginning of mention, $I$ for intermediate of mention, and $O$ for non-mention. As PLMs adopt different tokenization methods for words, we only annotate the first token of each word and fill in the rest using the special character $\left \langle pad \right \rangle$.

\subsubsection{Entity Disambiguation}\label{sec3.2.2}
The mention $m$ representing the entity name will be used to link to the grounded nodes in the KG. We pre-generate an inverted indexed dictionary which establishes a mapping of mentions to entities. We use $m$ to look up the corresponding KG entities in the inverted index dictionary, which are regarded as candidate entities $E_c$. For instance, we obtain a set of candidate entities according to the mention ``$washington$'', including the capital of the United States ``\textit{Washington, D.C.}'', the state ``\textit{Washington}'' and the person ``\textit{George Washington}''. Besides, the adjacent relations $r_{e_i}$ for each $e_i\in E_c$ retrieved from KG will be used for disambiguation. For example, the adjacent relations of the person ``\textit{George Washington}" are ``$born\_in$'', ``$died\_in$'', ``$founded\_organisation$", etc.

Various PLMs are employed to score the entity $e_i\in E_c$, and the formula is $S_{e_i}=\mathtt{g_{PLM}}(Q\vert e_i \vert r_{e_i})$, where $\vert$ refers to the connection symbol, $\mathtt{g_{PLM}}()$ represents an PLM encoder. The loss function of the entity disambiguation model based on PLM is:

\begin{equation}
\mathcal{L_{ED}} = -log P(y=e_k) ,
\end{equation}

\begin{equation}
P(y=e_k) = \frac{e^{S_{e_k}}}{e^{S_{e_k}}+\sum_{j=1}^{N}e^{S_{e_j}}},
\end{equation}
where $e_k$ denotes the gold entity, $N$ indicates the number of negative samples, $e_j$ represents negative entities and $P(y$=$e_k)$ is the probability of $e_k$. In addition, we adopt a simple linguistic approach, fuzzy string matching, to initially rank $E_c$ to select more challenging negative sample entities to train the model. We initial rank $E_c$ according to the Levenshtein Distance score between the entity name and the mention $m$.

The entity set $E_c$ with scores is obtained at the inference stage.


\subsubsection{Relation Detection}
Relation detection is a PLMs-based classification task in this framework. Since simple KGQA only considers one-hop relations, a question $Q$ corresponds to only one relation in KG. The model aims to map $Q$ to a KG relation $r\in \mathbb{R}$. For instance, the question ``\textit{where is washington locate}" in Fig.\ref{CLFramework} corresponds to the relation ``\textit{located\_in}" in KG.
Specifically, PLMs are employed to encode the question sequence to obtain the vector $h$, and then $h$ is fed into the linear classification layer to obtain the probability distribution of relations\footnote{The dimension of $h$ is $1\times1$. Different PLMs obtain $h$ in different ways, e.g.  $h = w\cdot h_{\left[ CLS\right]}^{T}$ in BERT.}. The goal of the model is to minimize:

\begin{equation}
\mathcal{L_{RD}} = -log P(y=\hat{r}\vert Q) ,
\end{equation}

\begin{equation}
P(y=\hat{r}\vert Q) = \frac{e^{h_{\hat{r}}}}{\sum_{j=1}^{M}e^{h_{r_j}}},
\end{equation}
where $\hat{r}$ refers to the gold relation, $P(y=\hat{r}\vert Q)$ represents the probability of $\hat{r}$ and $M$ denotes to the number of relation categories. Finally, the relation set $R_c$ with scores is obtained.

\subsubsection{Answer Query}
This module does not involve any neural networks and aims to query the answer in KG using entity-relation pair.
Given the set of candidate entities $E_c$ and the set of candidate relations $R_c$ obtained by the entity disambiguation and relation detection modules, we combine them into $(e, r)$ pairs to be queried in KB, where $e \in E_c$ and $r \in R_a$. We rank each $(e, r)$ pair, whose score is the weighted sum of its component scores, i.e., the entity disambiguation score and the relation detection score. The score of the $(e, r)$ pair is

\begin{equation}
S_{(e, r)} = \lambda S_e + (1-\lambda)S_r ,
\end{equation}
where $\lambda \in (0, 1)$, tuned according to the result of validation set. $S_e$ and $S_r$ are normalized entity score and relation score, respectively.

Note that the $(e, r)$ pair may be invalid because such a combination does not exist in KG. We remove these pairs by querying and verifying them in KG. In addition, the popularity of entities is applied to further prune pairs for the same score. In our work, the popularity is derived from FACC1\footnote{http://lemurproject.org/clueweb12/FACC1/} and the degree of entities.

\subsection{The Basic Retrieval and Ranking-based KGQA Framework} 
\begin{figure}
\begin{center}{
\includegraphics[width=0.85\textwidth]{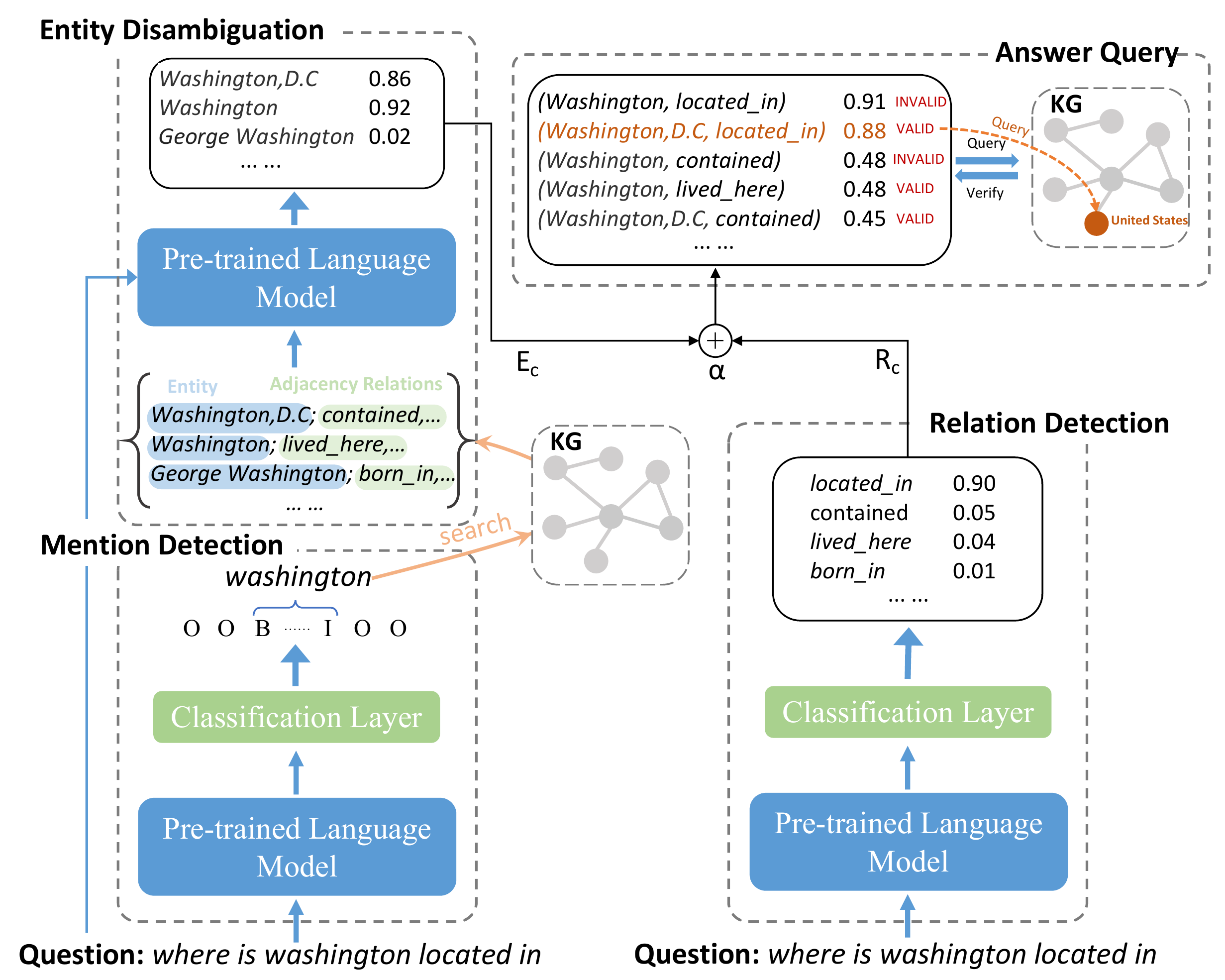}
\caption{The basic retrieval and ranking-based KGQA framework.} \label{RRFramework}
}
\end{center}
\end{figure}

The \textbf{R}etrieval and \textbf{R}anking-based \textbf{KGQA} framework (\textsc{KGQArr}) is shown in Fig.\ref{RRFramework}. It is a pipeline structure and consists of four modules, of which mention detection, entity disambiguation and answer query are identical to \textsc{KGQAcl}, differing only in the relation detection module.

Different from \textsc{KGQAcl}, the aim of relation detection in \textsc{KGQArr} is to select the relation in candidate relations $R_c$ that has the highest semantic similarity score to the question pattern $p$. $R_c$ consists of all adjacent relations searched by candidate entity $E_c$ in KG. The question pattern $p$ is obtained by using a special token $\left \langle e \right \rangle$ by mention $m$ to replace the mention $m$ in the question $q$,

Following the way of Sentience-Bert \cite{ref_article28}, we employ two PLMs that share parameters to encode questions pattern and relations, respectively. This way of encoding significantly improves efficiency compared to cross-encoding.
For each relation $r_i \in R_c$, we compute their similar score $Score(p, r_i)$. The final predicted relation $\hat{r}$ is given by the following formula:

\begin{equation}
\hat{r}= \mathrm{argmax}_{r_{i}\in R_{c}}Score(p,r_{i}),
\end{equation}

\begin{equation}
Score(q,r_{i})=\mathrm{cos}(Pool(h_p),Pool(h_{r_i})),
\end{equation}
where $h_p$ and $h_{r_i}$ are both obtained by PLMs, $Pool()$ refers to the pooling layer. During training, we adopt the hinge loss to maximize the margin between the gold relation $r^+$ and the negative relation $r^-$ in $E_c$.

\begin{equation}
\mathcal{L_{RD}}=\sum_{i=1}^{k}\mathtt{max}\left \{ 0,\gamma -Score(p,r^+)+Score(p,r_{i}^-) \right \},
\end{equation}
where $\gamma$ is a constant parameter, where $k$ is the number of negative relations. As with the \textsc{KGQAcl} framework, the candidate relations $R_c$ with scores and the candidate entities $E_c$ with scores will be fed into the answer query module together.

\section{Benchmarks}\label{Sec4}

In this section, we will describe the four benchmarks utilized for the experiments and the method for constructing the benchmarks (section \ref{ConstructeBenchmarks}). In addition, we introduce accuracy and efficiency evaluation metrics, as well as the method for evaluating the scalability of the KGQA system based on PLMs (section \ref{EvaluationMetrics}).

\subsection{Construction of the benchmarks}\label{ConstructeBenchmarks}

We construct experiments on four benchmarks. Apart from the popular simple KGQA benchmark SimpleQuestions \cite{ref_article25}, we construct three more benchmarks to explore the scalability of PLMs on KGQA. In particular, we increase the scale of the original KG of SimpleQuestion and propose the three KGQA benchmarks to investigate the performance changes of PLMs as the KG size increases. Note that the question-answering datasets of the four benchmarks are the same.

\textbf{(a)} The original SimpleQuestions with small-scale KG (\textsc{SQs}) \cite{ref_article25}. The original benchmark contains more than 100,000 questions, divided into train/validation/test on a 7/1/2 split. The KG resource of this benchmark is FB2M, denoted as $\mathbb{G}_{S}$, which contains 2M entities and 6.7K relations. Some previous works pre-pruned KB to fit their methods because they assumed all questions were known. We do not preprocess KG to do experiments with all comparison models. 

\textbf{(b)} SimpleQuestions with large-scale KG (\textsc{SQl}). Benchmark \textsc{SQl} requires getting a triple in a large-scale KG to answer questions. For the construction of the KG of \textsc{SQl}, denoted as $\mathbb{G}_{L}$, we retrieve all one-hop triples of the entities in FACC1 in the Freebase dump\footnote{https://developers.google.com/freebase}. FACC1 provides the common names and the popularity of the entities. We then merge $\mathbb{G}_{FACC1}$ with $\mathbb{G}_{S}$ to obtain the KG $\mathbb{G}_L$: $\mathbb{G}_L = \mathbb{G}_{FACC1} \bigcup \mathbb{G}_{S}$. $\mathbb{G}_L$ contains 108M entities, 12.7K relations and 292M triples, which completely covers $\mathbb{G}_{S}$. 

Additionally, we construct two more benchmarks with KGs as $\mathbb{G}_{M-A}$ and $\mathbb{G}_{M-B}$ respectively. Their number of triples is between that of $\mathbb{G}_{S}$ and $\mathbb{G}_{L}$, and the number of triples of these four KGs grows uniformly.

\textbf{(c)} SimpleQuestions with medium-scale KG$_A$ (\textsc{SQm-a}).  The KG $\mathbb{G}_{M-A}$ of \textsc{SQm-a} includes 61M entities, 11.1K relations and 105M triples, which also completely covers $\mathbb{G}_{S}$.

\textbf{(d)} SimpleQuestions with medium-scale KG$_B$ (\textsc{SQm-b}).  The KG $\mathbb{G}_{M-B}$ of \textsc{SQm-b} includes 90M entities, 12.2K relations and 202M triples, which also completely covers $\mathbb{G}_{S}$.

The overall comparison of the KG for the four benchmarks is shown in Table \ref{benchmarksTable}. Apart from the number of entities, relations and triples for the KG of each benchmark, we also count the average degree, i.e., the average number of adjacent relations of the entities appearing in the SimpleQuestions dataset. Average degree can reflect the challenge of the benchmark to some extent.

\begin{table}[]
\centering
\caption{The overall comparison of the KG for the four benchmarks. Average Degree represents the average number of adjacent relations of the entities appearing in the dataset.}\label{benchmarksTable}
\begin{tabular}{|l|c|c|c|c|}
\hline
      & Entities & Relations & Triples & Average Degree \\ \hline
\textsc{SQs}  & 1,951,909 & 6,701     & 7,188,636       & 88             \\ \hline
\textsc{SQm-a} & 61,558,084       & 11,127         & 105,003,132       & 176            \\ \hline
\textsc{SQm-b} & 90,282,560        & 12,227         & 202,002,048      & 293            \\ \hline
\textsc{SQl}   & 108,183,882 & 12,794    & 292,088,464       & 407            \\ \hline
\end{tabular}
\end{table}

\subsection{Evaluation Metrics}\label{EvaluationMetrics}
We evaluate the overall performance of KGQA in terms of \textbf{accuracy} and \textbf{efficiency}.
The accuracy metric follows the common evaluation method for SimpleQuestions, where we calculate the accuracy of inferred ($s, r$) pairs. Only a fact matching the ground truth answer in both subject $\hat{s}$ and predicate $\hat{r}$ is correct.

\begin{equation}
accuracy= \frac{\sum_{i=1}^N1_{[(\hat{s}_i,\hat{r}_i)=(s_i,r_i)]}}{N},
\end{equation}
where $N$ refers to the number of questions.

We calculate the average training and test time of all KGQA systems to evaluate their efficiency. For a fair comparison, we set the same batch size and negative sampling number for each PLM in the same basic framework.

Additionally, we defined the Variation in Accuracy (VA) and the Variation in average test Time (VT) to evaluate the \textbf{scalability} of the KGQA system.

\begin{definition}[VA]
VA of a KGQA system in a benchmark KG$_x$ represents the gap between the accuracy of the system on KG$_x$ and the accuracy of the system on small-scale KG benchmarks \textsc{KGs}.
$\mathrm{VA} = accuracy_{\textsc{KGs}}-accuracy_{KG_x}$, where KG$_x$ represents one of our four benchmarks. A \textbf{higher} VA means that the system performs worse in terms of scalability.
\end{definition}

\begin{definition}[VT]
VT of a KGQA system on a benchmark KG$_x$ represents the change in the average test time of the system on KG$_x$ compared to the average test time of the system on small-scale KG benchmarks \textsc{KGs}. $\mathrm{VT} = time_{KGx}-time_{\textsc{KGs}}$, where KG$_x$ represents one of our four benchmarks.
A \textbf{higher} VT means that the system performs worse in terms of scalability.
\end{definition}

\section{Experiments}\label{Sec5}

In this section, we first present all the PLMs-based KGQA systems (Section \ref{AllSystems}) and the experimental setup (Section \ref{ExperimentalSetup}). We then show the overall experimental results and discuss the results in light of the three research questions (section \ref{OverallResultsDiscussions}). We further explore the sub-modules of KGQA and discuss them according to the three new research questions (Section \ref{sec4.3}). Besides, we also evaluate all systems on two other KGQA datasets apart from the SimpleQuestions family (Section \ref{Sec6.5}).
Finally, we compare the performance between ChatGPT and other PLMs on datasets SimpleQuestions, WebQuestionSP and FreebaseQA (Section \ref{Sec6.6}).

\subsection{All KGQA Systems Based on PLMs}\label{AllSystems}
In this work, 18 KGQA systems (9 PLMs * 2 basic KGQA frameworks) were implemented for evaluation. 
\textbf{Three} classes of \textbf{nine} PLMs were used for evaluation, including the common large-scale PLMs \textsc{Bert}\footnote{https://huggingface.co/bert-base-uncased.}, \textsc{Roberta}\footnote{https://huggingface.co/roberta-base.}, \textsc{XLnet}\footnote{https://huggingface.co/xlnet-base-cased.}
and \textsc{Gpt2}\footnote{https://huggingface.co/gpt2.} , the lightweight PLMs \textsc{ALbert}\footnote{https://huggingface.co/albert-base-v2.}, \textsc{DistilBert}\footnote{https://huggingface.co/distilbert-base-uncased.} and \textsc{DistilRoberta}\footnote{https://huggingface.co/distilroberta-base.}, and the knowledge-enhanced PLMs \textsc{Luke}\footnote{https://huggingface.co/studio-ousia/luke-base.} and \textsc{Kepler}\footnote{https://github.com/THU-KEG/KEPLER.}. 
As common models serve as the backbone models for lightweight PLMs and knowledge-enhanced PLMs, we follow \cite{Pre-trainedModels} to further classify the common PLMs according to pre-trained task categories, namely, Masked Language Modeling (MLM, i.e., \textsc{Bert} and \textsc{Roberta}), Language Modeling (LM, i.e., \textsc{Gpt2}), and Permuted Language Modeling (PeLM, i.e., \textsc{XLnet}).
Parameters of these PLMs\footnote{These data are from https://huggingface.co/.} are shown in Table \ref{ParametersOfPLMs}.

\begin{table}[]
\centering
\caption{Parameters of various PLMs.}\label{ParametersOfPLMs}
\scalebox{0.75}{
\begin{tabular}{|l|c|c|c|c|c|c|c|c|c|}
\hline
PLMs           & \textsc{Bert} & \textsc{Roberta} & \textsc{XLnet} & \textsc{Gpt2} & \textsc{ALbert} & \textsc{DistilBert} & \textsc{DistilRoberta} & \textsc{LUke} & \textsc{KEpler}  \\ \hline
Parameters  & 110M                           & 125M                              & 117M                            & 125M                           & 11M                              & 66M                                  & 82M                                     & 253M                           & 125M \\ \hline
\end{tabular}
}
\end{table}


\textbf{BERT}. BERT is the most representative pre-trained language model that uses the encoder of the deep Transformer as its backbone. BERT uses Masked Language Modelling (MLM) and Next Sentence Prediction (NSP) as self-supervised tasks for pretraining.

\textbf{RoBERTa}. RoBERTa has almost the same architecture as BERT, while it differs in the parameter settings and training objectives. RoBERTa removes the NSP loss and creates the dynamic MLM mask instead of the static mask used in BERT to train a larger scale and a longer sequence model.

\textbf{GPT2}. Unlike BERT and Roberta, which are all masked language models, GPT2 is an autoregressive language model predicting one token at a time from left to right (i.e. LM). GPT2 is often used for natural language generation, whereas BERT and Roberta are mainly used for natural language understanding.

\textbf{XLNET}. XLNET is known as an permuted language model \cite{Pre-trainedModels}. Unlike GPT2 can not utilize the context from the backward side, XLNET resolves this problem by adopting a new objective called Permutation Language Modeling (PeLM), enabling the model to take advantage of both forward and backward contexts.

\textbf{ALBERT}. ALBERT is a lite version of BERT. All its Transformer blocks share parameters and its embedding matrix is decomposed into two smaller matrices. Thus ALBERT has a much smaller number of parameters than BERT. Instead of NSP, Albert predicts the order of two consecutive text segments.

\textbf{DistilBERT}. DistilBERT is a distilled version of BERT that is pre-trained on the same corpus in a self-supervised manner, using the BERT model as a teacher. This means that it only pre-trains on raw texts, with no humans labeling them in any way.

\textbf{DistilRoBERTa}. DistilRoBERTa is a distilled version of RoBERTa. It follows the same training procedure as DistilBERT.

\textbf{LUKE}. LUKE is based on RoBERTa and adds entity embeddings as well as an entity-aware self-attention mechanism. The entity-aware self-attention mechanism is an extension of the self-attention mechanism of the Transformer and considers the types of words or entities when computing attention scores.

\textbf{KEPLER}. KEPLER is a unified model for knowledge embedding (KE) and PLM representation. It encodes textual entity descriptions with a PLM and then optimizes KE and language modeling objectives jointly.

\subsection{Experimental Setup}\label{ExperimentalSetup}
All KGQA systems were trained using an NVIDIA GeForce RTX 2080 TI. We performed a grid search for all KGQA systems, choosing the hyperparameter configuration that achieves the highest final accuracy. 
We adopted an early stop strategy in training and set the patience to 3. Because the PLMs-based entity disambiguation model is too large, the batch size can only be set to 1. Nevertheless , this may cause some PLMs to be difficult to converge, so we used the gradient accumulation method to increase the gradient accumulation step instead of increasing the batch size. Note that the value of the batch size multiplied by the gradient accumulation step is the same for all PLMs in the same sub-module task to ensure a fair comparison of their training times. In addition, KGQA systems output Top-50 results for entity disambiguation and Top-5 results for relation detection to combine (subject, relation) pairs.

\subsection{Overall Results and Discussions}\label{OverallResultsDiscussions}
In this section, we show the overall results of the 18 KGQA systems based on various PLMs in terms of accuracy and efficiency (i.e., average training time and average testing time). Based on these experimental results, we will discuss three questions: (1) What PLMs have the best accuracy or efficiency performance? (2) What are the differences in accuracy and efficiency between the two basic KGQA frameworks? (3) How scalable are the various PLMs, i.e., how do their accuracy and efficiency vary as the size of KG increases? 

\subsubsection{Discussion on the Accuracy of KGQA systems}\label{sec5.3.1}
The accuracy performance of all the studied PLMs-based KGQA frameworks and benchmarks are summarised in Table \ref{AccOfALl}. 

The bold numbers in Table \ref{AccOfALl} represent the highest accuracy of all KGQA system results using one benchmark.
\textsc{Roberta} and the two knowledge-enhanced PLMs \textsc{Luke} and \textsc{Kepler} achieves the best accuracy results. 
Both \textsc{Luke} and \textsc{Kepler} are based on \textsc{Roberta} for knowledge augmentation. The results demonstrate the powerful performance of \textsc{Roberta} and that knowledge enhancement is beneficial for knowledge-intensive tasks, i.e., KGQA.
\textsc{Luke} and \textsc{Kepler} perform better on small-scale KG benchmarks while \textsc{Roberta} perform better on large-scale KG benchmarks. This may be due to the fact that the introduction of additional knowledge-enhancing pre-training objects affects the robustness of the model itself.
Comparing the performance of the same PLM on the two basic frameworks, we found that the \textsc{KGQArr} framework significantly outperforms the \textsc{KGQAcl} framework on small-scale KG benchmark \textsc{SQs}. However, as the size of KG increases, the accuracy of \textsc{KGQArr} becomes inferior to that of \textsc{KGQAcl}. We will investigate the reason for this in Section \ref{sec4.3} by analyzing the performance variation of their submodules.
In addition, we noted that \textsc{XLnet} and \textsc{Gpt2} are worse than the other PLMs in terms of accuracy in almost all settings, even for the three lightweight PLMs. In particular, the \textsc{Gpt2}-based \textsc{KGQArr} system performed extremely poorly in all benchmarks. We believe that the modelling way of PLMs influences it. All PLMs are modelled by the auto-encoding way (i.e. MLM) except \textsc{XLnet} and \textsc{Gpt2}. 
\textsc{Gpt2} is modeled via auto-regressive way (i.e. LM), while \textsc{XLnet} combines the idea of auto-encoding with auto-regressive modeling ((i.e. PeLM)).
Auto-encoding modelling is skilled in solving natural language understanding (NLU) tasks, while auto-regressive modelling is proficient in solving natural language generation (NLG) tasks.
Therefore, \textsc{XLnet} and \textsc{Gpt2} perform poorly on KGQA because traditional approaches treat it as an NLU task to solve.
There has been some recent work \cite{unifiedSKG} to convert KGQA to the NLG task for handling, and we will explore this approach in the future.

\begin{table}[hbp]
 \centering
 \renewcommand\tabcolsep{1pt}
 \renewcommand{\arraystretch}{1.2}
 \caption{Overall accuracy (\%) of different PLMs-based KGQA systems on four benchmarks. Bolded numbers indicate the highest accuracy.}\label{AccOfALl}
 \begin{tabular}{lc|cccc|cccc}
  \toprule
   \multicolumn{2}{c|}{\textbf{Frameworks}}&  \multicolumn{4}{|c|}{\textsc{KGQAcl}} & \multicolumn{4}{|c}{\textsc{KGQArr}}\\
  \midrule
   \multicolumn{2}{c|} {\diagbox{\textbf{PLMs}}{\textbf{Benckmarks}} }&  \textsc{SQs}  & \textsc{SQm-a} & \textsc{SQm-b} & \textsc{SQl}&\textsc{SQs} & \textsc{SQm-a}& \textsc{SQm-b}& \textsc{SQl} \\
  \midrule
  \multicolumn{10}{c}{\textit{Common PLMs}} \\
  \midrule
\textsc{Bert} (\textit{MLM)} & &74.79 &68.02 &66.03&64.81 &76.76 &67.75&66.54&65.39\\
  \midrule
\textsc{Roberta} (\textit{MLM)} & &75.32&68.97&\textbf{66.33}&\textbf{65.63}&77.12&68.20&\textbf{66.81}&\textbf{65.66}\\
  \midrule
\textsc{XLnet} (\textit{PeLM)} & &74.30&66.47&65.31&59.92&76.26&66.41&65.40&58.55\\
  \midrule
\textsc{Gpt2} (\textit{LM)}& &73.39&66.52&64.53&64.30&23.18&20.04&19.88&19.75\\
 \midrule
 \multicolumn{10}{c}{\textit{Lightweight PLMs}} \\
  \midrule
\textsc{ALbert} & &74.26&67.01&65.37&64.66&76.20&66.80&65.60&65.10\\
   \midrule
\textsc{DistilBert} & &74.29&67.87&65.45&64.36&75.11&68.07&65.55&64.46\\
   \midrule
\textsc{DistilRoberta} & &74.04&67.32&64.67&64.39&73.24&67.37&65.48&64.00\\
   \midrule
\multicolumn{10}{c}{\textit{Knowledge-enhanced PLMs}} \\
  \midrule
\textsc{Luke} & &\textbf{75.67}&\textbf{69.25}&65.26&64.86&77.00&67.97&65.11&64.26\\
   \midrule
\textsc{Kepler} &  &74.87&68.41&65.23&65.15&\textbf{77.41}&\textbf{68.41}&66.20&64.94\\
  \bottomrule
 \end{tabular}
\end{table}

\begin{figure}
\begin{center}{
\includegraphics[width=0.95\textwidth]{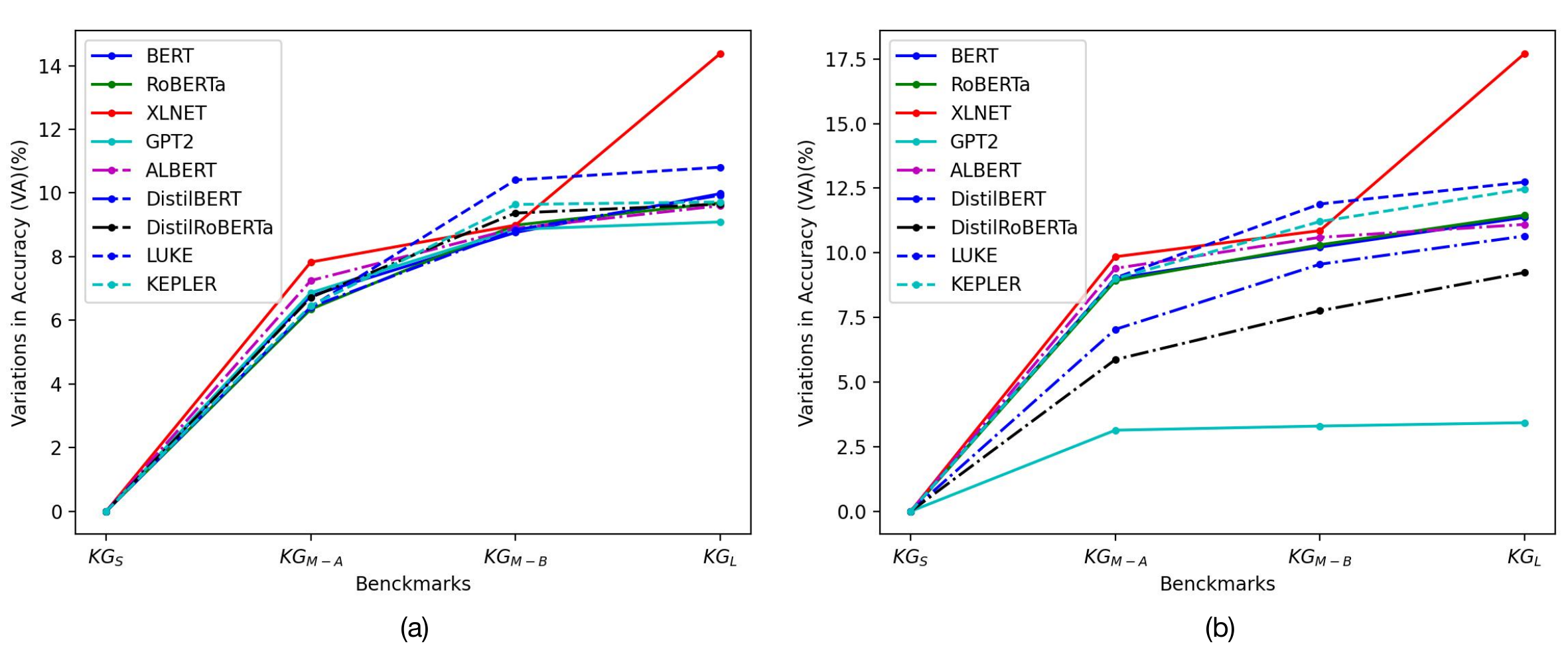}
\caption{Scalability of all \textsc{KGQAcl} frameworks (a) and \textsc{KGQArr} frameworks (b) in terms of accuracy variation.} \label{VAofTwoFrameworks}
}
\end{center}
\end{figure}

We investigated the scalability of the KGQA system with two metrics, VA and VT, defined in Section \ref{EvaluationMetrics}.
As shown in Fig.\ref{VAofTwoFrameworks}, the VA of all PLMs-based KGQA systems under both basic frameworks shows an increasing trend, indicating that scalability gradually worsens as KG size increases. We excluded the analysis of \textsc{Gpt2} due to its terrible accuracy. Among the KGQAcl and KGQArr frameworks, XLnet exhibits the worst scalability performance, especially for the benchmark \textsc{SQl} with the largest KG size. 
In addition, the knowledge-enhanced PLMs \textsc{Luke} and \textsc{Kelpler} perform inferiorly to other PLMs in terms of scalability on the larger scale KG benchmarks (\textsc{SQm-b} and \textsc{SQl}). In contrast, the lightweight PLMs \textsc{ALbert}, \textsc{Distilbert} and \textsc{Distilrobert} are more robust to KG scale variations and perform better in terms of scalability. We will further analyze which sub-modules in the framework primarily affect scalability in section \ref{sec4.3}.

\begin{figure}
\begin{center}{
\includegraphics[width=0.95\textwidth]{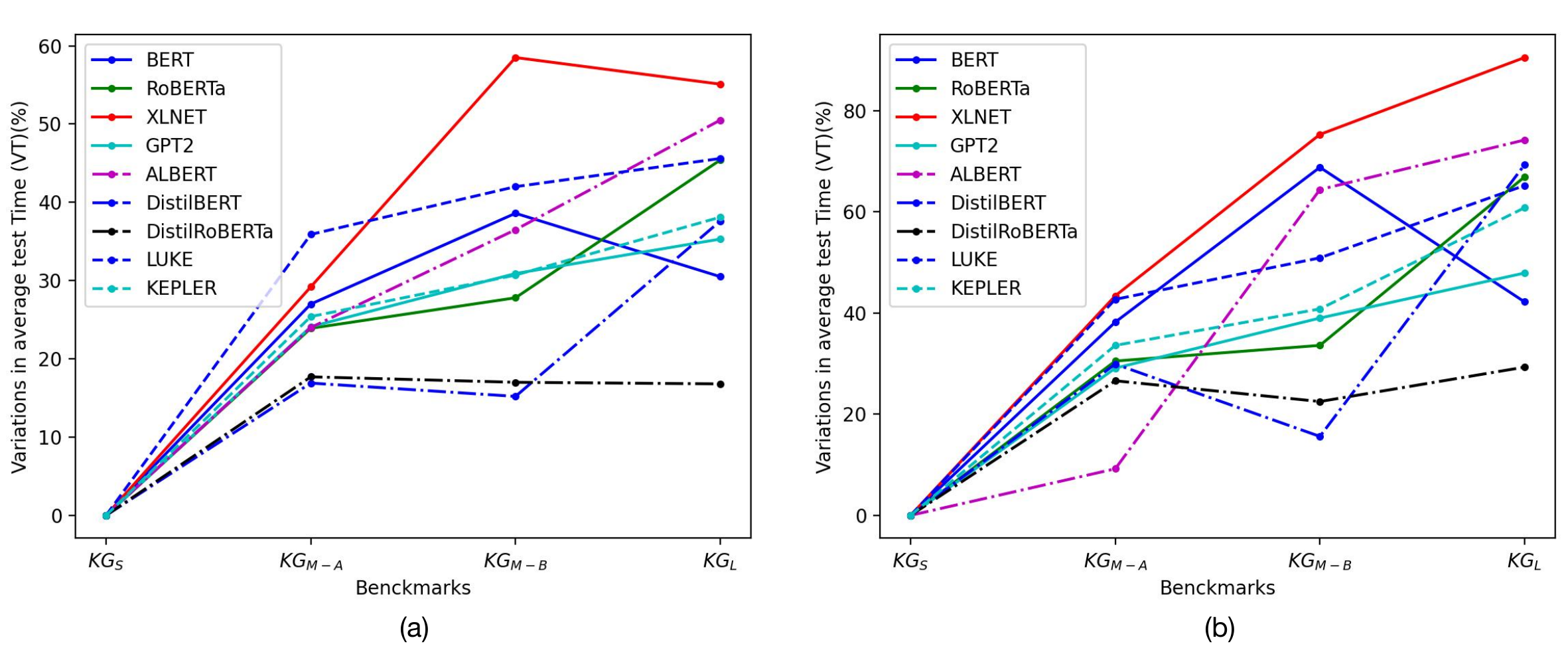}
\caption{Scalability of all \textsc{KGQAcl} frameworks (a) and \textsc{KGQArr} frameworks (b) in terms of average test time variation.} \label{VTofTwoFrameworks}
}
\end{center}
\end{figure}


\subsubsection{Discussion on the Efficiency of KGQA Systems}
The efficiency performance of all the studied PLMs-based KGQA frameworks and benchmarks are summarised in Table \ref{OverallEfficiency}.
We set the same patience for all KGQA systems so that each model was trained to converge. Due to the different convergence rates of the different PLMs, the variation of the training time of the PLMs did not coincide with the variation of the testing time.

As shown in Table \ref{OverallEfficiency}, the two lightweight PLMs \textsc{DistilBert} and \textsc{DistilRoberta} exhibit the highest efficiency in training and testing. \textsc{DistilBert} is up to 3.1x faster than \textsc{XLnet} on training (2325.5 ms vs. 749.7 ms), and \textsc{DistilBert} is up to 3.2x faster than \textsc{XLnet} on testing (85.3 ms vs. 26.3 ms).
Another lightweight model, \textsc{ALbert}, has the least number of parameters, but its efficiency does not have any advantage over other PLMs. Therefore, the knowledge distillation approach is an effective way to improve efficiency, while the matrix parameter sharing strategy only reduces GPU memory consumption, with no improvement in efficiency.
In addition, the time consumption of all PLMs tends to increase as the size of KG increases.
Comparing the efficiency of different basic frameworks for the same PLM, \textsc{KGQArr} is always more time-consuming than \textsc{KGQAcl}. According to the analysis in Section \ref{sec5.3.1}, the \textsc{KGQArr} framework is only more accurate than the \textsc{KGQAcl} framework for small-scale KG benchmark \textsc{SQs}. Therefore, the \textsc{KGQAcl} framework is a better choice for large-scale KG benchmarks.

The efficiency scalability of all PLMs based on both frameworks is shown in Fig.\ref{VTofTwoFrameworks}. All PLMs show an increasing trend in VT as the KG size increases. Among them, \textsc{DistilRoberta} has the best scalability as it has the smallest VT on all benchmarks. Another lightweight PLM, \textsc{DitilBert}, also shows good scalability. Section \ref{sec5.3.1} also demonstrates \textsc{DistilRoberta} and \textsc{DitilBert} have the equivalent accuracy performance as the other PLMs on large-scale KG. These findings indicate that the two knowledge distillation PLMs have excellent scalability. The knowledge distillation is a promising approach for PLMs applied to KGQA.

\begin{table}[hbp]
 \centering
 \renewcommand\tabcolsep{1pt}
 \renewcommand{\arraystretch}{1.2}
 \caption{Efficiency of all PLMs-based KGQA frameworks on four benchmarks. Tr denotes average training time (ms) and Te denotes average test time (ms). Underlined ones indicate the shortest time spent.}\label{OverallEfficiency}
 \begin{tabular}{lc|cccc|cccc}
  \toprule
   \multicolumn{2}{c|}{\textbf{Frameworks}}&  \multicolumn{4}{|c|}{\textsc{KGQAcl}} & \multicolumn{4}{|c}{\textsc{KGQArr}}\\
  \midrule
   \multicolumn{2}{c|} {\diagbox{\textbf{PLMs}}{\textbf{Benckmarks}} }&  \textsc{SQs}  & \textsc{SQm-a} & \textsc{SQm-b} & \textsc{SQl}&\textsc{SQs} & \textsc{SQm-a}& \textsc{SQm-b}& \textsc{SQl} \\
  \midrule
  \multicolumn{10}{c}{\textit{Common PLMs}} \\
  \midrule
\textsc{Bert} (\textit{MLM)} &Tr &489.1 &492.5 &1042.8 &584.5&1296.8&1190.4&2538.0&1563.7\\
&Te &45.9 &72.9 &84.5&76.4&70.4&108.6&139.2&112.6\\
  \midrule
\textsc{Roberta} (\textit{MLM)}&Tr &523.6 &782.8 &701.0&772.6&1450.1&1341.8&1470.2&1906.1\\
&Te &45.8 &69.7 &73.6 &91.2&70.0&100.5&103.6&136.9\\
  \midrule
\textsc{XLnet} (\textit{PeLM)}&Tr &850.4 &602.7 &631.3&655.6&2325.5&2500.7&2308.9&1685.3\\
&Te &85.3 &114.5 &143.8&140.4&117.5&160.9&192.8&208.0\\
  \midrule
\textsc{Gpt2} (\textit{LM)}&Tr &467.2 &879.6 &484.9&502.1&854.2&1341.1&1078.0&1076.9\\
&Te &48.4 &72.5 &79.3&83.7&75.6&104.7&114.6&123.5\\
 \midrule
   \multicolumn{10}{c}{\textit{Lightweight PLMs}} \\
   \midrule
\textsc{ALbert}&Tr &661.8 &709.2 &581.8&896.6&2679.6&1609.0&1500.8&1969.8\\
&Te &54.9 &78.9 &91.4&105.4&105.6&114.8&170.0&179.8\\
   \midrule
\textsc{DistilBert} &Tr &\underline{452.3} &\underline{521.6}&\underline{553.2}&\underline{463.3}&\underline{749.7}&\underline{970.7}&\underline{1169.4}&\underline{1120.8}\\
&Te &\underline{26.3} &\underline{43.2} &\underline{41.5}&63.9&48.4&78.3&\underline{64.0}&117.7\\
   \midrule
\textsc{DistilRoberta} &Tr &477.5 &902.6&871.4&901.0&801.7&1422.6&1504.1&1290.6\\
&Te &28.0 &45.7 &45.0&\underline{44.8}&\underline{42.9}&\underline{69.5}&65.4&\underline{72.2}\\
   \midrule
\multicolumn{10}{c}{\textit{Knowledge-enhanced PLMs}} \\
   \midrule
\textsc{Luke} &Tr &462.7 &1227.3&590.5&634.1&1738.6&2430.3&2213.9&1609.2\\
&Te &41.5 &77.4 &83.5&87.1&67.9&110.6&118.8&133.1\\
   \midrule
\textsc{Kepler} &Tr &960.5 &1299.1&753.4&1130.7&2164.4&2522.4&2100.8&2011.5\\
&Te &48.4 &73.8 &79.1&86.5&69.0&102.6&109.8&129.8\\
  \bottomrule
 \end{tabular}
\end{table}


\subsubsection{Summary and New Research Questions}
Some important conclusions can be drawn from the above discussion. \textsc{Roberta}, \textsc{Luke} and \textsc{Kepler} perform best in terms of overall accuracy. Nevertheless , Luke and Kepler have slightly poorer scalability, with a greater variation in accuracy as the KG size increases. The two lightweight PLMs \textsc{DistillBert} and \textsc{DistilRoberta} exhibit the best scalability in accuracy and efficiency. Their accuracy on large-scale KG is the same as other PLMs, and their inference time is up to 3.3x faster than other PLMs. For the KGQA framework, the \textsc{KGQArr} framework is significantly less efficient than the \textsc{KGQAcl} framework. Furthermore, the higher accuracy of \textsc{KGQArr}-based systems than \textsc{KGQAcl}-based systems is only at small KG scales. As the KG size increases, the \textsc{KGQAcl}-based system gradually outperforms the \textsc{KGQArr}-based system, which indicates the poor scalability of the \textsc{KGQArr} framework.

These findings lead us to explore the following questions further. (1) What sub-modules of the KGQA system are primarily responsible for the differences in accuracy and efficiency? (2) What sub-modules in the KGQA system are most susceptible to the variation in the size of KG? (3) Why does the \textsc{KGQArr} framework have worse scalability?
We explore these three questions by examining the performance of the submodules of all KGQA systems.

\subsection{Study of the KGQA Sub-modules}\label{sec4.3}
\subsubsection{Results and Discussion on KGQA Sub-modules}\label{sec4.3.1}
We further compared the sub-module performance of each KGQA system in this section to explore the primary influencers of accuracy and efficiency for each PLMs. Additionally, we compared the two KGQA base frameworks to explore the reasons for their large variability.

Tables \ref{MDResults}, \ref{EDResults} and \ref{RDResults} show the overall results for Mention Detection (MD), Entity Disambiguation (ED) and Relation Detection (RD) respectively. For efficiency, we only compare the average test time. We do not analyze Anwer Query further as it is irrelevant to PLMs. The final result of MD is not affected by the KGQA basic frameworks and benchmarks, and it is only relevant to PLMs. 
Table \ref{MDResults} shows that all PLMs except \textsc{Gpt2} have similar accuracy and efficiency on MD. It indicates that \textsc{Gpt2}, based on auto-regressive modelling (i.e. LM), is not good at solving NER tasks. Bert has the highest F1 value but poor efficiency. 
Notice that \textsc{Bert}'s distilled version \textsc{DistilBert} improves efficiency by almost double and has only a slight performance penalty.

The ED of both frameworks is the same. As shown in Table \ref{EDResults}, \textsc{RoBerta} exhibits the best accuracy performance, and \textsc{DistilBERT} and \textsc{DistilRoBERTa} have the shortest test time. It is worth to note that the accuracy and efficiency of all PLMs in the ED task are greatly affected by the KG size. This is because as the KG size increases, the number of candidate entities and the degree of entities increases, as shown in Table \ref{benchmarksTable}. But, as the KG size gets larger, the impact on accuracy becomes smaller. \textsc{XLnet} shows the most severe decrease in accuracy (27.59\% decrease), and \textsc{ALbert} shows the most significant increase in test time (50.3ms increase).

\begin{table}[hbp]
 \centering
 \renewcommand\tabcolsep{1pt}
 \renewcommand{\arraystretch}{1.2}
 \caption{Results of Mention Detection of all PLMs-based KGQA systems. Te refers to the average test time (ms). Bolded numbers indicate the best performance and underlined ones indicate the shortest time spent.}\label{MDResults}
 \scalebox{0.9}{
 \begin{tabular}{l|cccc}
  \toprule
   \textbf{PLMs} & \textbf{Precision} & \textbf{Recall}& \textbf{F1} & \textbf{Te} \\
  \midrule
    \multicolumn{5}{c}{\textit{Common PLMs}} \\
  \midrule
\textsc{Bert} (\textit{MLM)}  &\textbf{95.12}&95.50&\textbf{95.31}&2.50\\
\midrule

\textsc{Roberta} (\textit{MLM)} &94.62&\textbf{95.77}&95.21&2.27\\
\midrule

\textsc{XLnet} (\textit{PeLM)} &95.05&95.28&95.16&4.00\\
\midrule

\textsc{Gpt2} (\textit{LM)} &85.85&88.15&86.98&\underline{0.73}\\
\midrule

  \multicolumn{5}{c}{\textit{Lightweight PLMs}} \\
  \midrule
\textsc{ALbert} &94.32&94.68&94.51&2.43\\
\midrule

\textsc{DistilBert} &94.58&95.37&94.97&1.15\\
\midrule

\textsc{DistilRoberta} &94.41&94.92&94.66&1.18\\
\midrule

  \multicolumn{5}{c}{\textit{Knowledge-enhanced PLMs}} \\
  \midrule
\textsc{Luke} &94.02&94.40&94.21&1.22\\
\midrule
\textsc{Kepler} &94.60&95.01&94.80&2.63\\
  \bottomrule
 \end{tabular}
 }
  \end{table}

\begin{table}[hbp]
 \centering
 \renewcommand\tabcolsep{1pt}
 \renewcommand{\arraystretch}{1.2}
 \caption{Results of Entity Disambiguation of all PLMs-based KGQA systems on four benchmarks. Top@1 denotes top-1 recall. Te refers to the average test time (ms). Bolded numbers indicate the best performance and underlined ones indicate the shortest time spent.}\label{EDResults}
 \scalebox{0.95}{
 \begin{tabular}{lc|cccc}
  \toprule
  \multicolumn{2}{l|}{\textbf{PLMs}} & \multicolumn{4}{c}{\textbf{Benchmarks}}\\
& & \textsc{SQs}& \textsc{SQm-a}& \textsc{SQm-b}& \textsc{SQl} \\
  \midrule
    \multicolumn{5}{c}{\textit{Common PLMs}} \\
  \midrule
\textsc{Bert} (\textit{MLM)} & Top@1 &82.26&73.58&72.79&71.39 \\
&Te&41.8&67.6&80.2&72.1\\
\midrule

\textsc{Roberta} (\textit{MLM)}& Top@1 &\textbf{82.57}&\textbf{75.32}&\textbf{73.85}&\textbf{71.80}\\
&Te&42.0&65.8&69.3&87.3\\
\midrule

\textsc{XLnet} (\textit{PeLM)} & Top@1 &82.46&69.12&67.75&54.87\\
&Te&78.5&107.6&136.8&87.3\\
\midrule

\textsc{Gpt2} (\textit{LM)}& Top@1 &80.17&71.49&69.4&69.18 \\
&Te&46.5&70.4&77.0&81.5\\
\midrule

  \multicolumn{5}{c}{\textit{Lightweight PLMs}} \\
  \midrule
\textsc{ALbert} & Top@1 &82.0&73.21&71.3&71.43\\
&Te&50.6&74.3&86.9&100.9\\
\midrule

\textsc{DistilBert} & Top@1 &81.80&72.59&70.41&67.7\\
&Te&\underline{24.3}&\underline{41.1}&\underline{39.0}&61.6\\
\midrule

\textsc{DistilRoberta} & Top@1 &81.59&72.75&68.76&69.14\\
&Te&26.0&44.5&42.7&\underline{42.6}\\
\midrule

  \multicolumn{5}{c}{\textit{Knowledge-enhanced PLMs}} \\
  \midrule
\textsc{Luke} & Top@1 &81.85&74.27&67.45&69.08\\
&Te&38.9&74.5&80.6&83.9\\
\midrule
\textsc{Kepler} & Top@1 &82.53&74.43&69.59&71.33 \\
&Te&44.0&69.2&74.5&81.6\\
  \bottomrule
 \end{tabular}
 }
  \end{table}

\begin{table}
 \centering
 \renewcommand\tabcolsep{1pt}
 \renewcommand{\arraystretch}{1.2}
 \caption{Results of Relation Dectection of all PLMs-based KGQA systems on four benchmarks. Top@1 denotes top-1 recall. Te refers to the average test time (ms).}\label{RDResults}
 \scalebox{0.95}{
 \begin{tabular}{lc|cccc|cccc}
  \toprule
   \multicolumn{2}{c|}{\textbf{Frameworks}}&  \multicolumn{4}{|c|}{\textsc{KGQAcl}} & \multicolumn{4}{|c}{\textsc{KGQArr}}\\
  \midrule
   \multicolumn{2}{c|} {\diagbox{\textbf{PLMs}}{\textbf{Benckmarks}} }& \textsc{SQs} & \textsc{SQm-a}& \textsc{SQm-b}& \textsc{SQl}&\textsc{SQs} & \textsc{SQm-a}& \textsc{SQm-b}& \textsc{SQl} \\
  \midrule
      \multicolumn{10}{c}{\textit{Common PLMs}} \\
  \midrule
\textsc{Bert} (\textit{MLM)} &Top@1 &81.22&80.79&80.50&80.43&85.14&81.86&81.69&81.02\\
&Te&1.5&1.6&1.5&1.5&26.1&33.4&56.3&37.8\\
\midrule

\textsc{Roberta} (\textit{MLM)}&Top@1&81.38&81.40&81.40&81.16&84.84&82.35&81.41&81.07\\
&Te&1.4&1.6&1.4&1.5&25.7&32.2&30.4&47.1\\
\midrule

\textsc{XLnet} (\textit{PeLM)} &Top@1&81.26&81.00&81.07&80.83&85.11&82.88&81.68&80.52\\
&Te&2.7&2.7&2.8&1.5&35.0&49.2&51.2&70.4\\
\midrule

\textsc{Gpt2} (\textit{LM)} &Top@1 &81.19&81.13&81.10&81.70&17.77&16.24&15.92&15.09\\
&Te&1.1&1.3&1.3&1.5&28.3&33.6&34.1&37.2\\
\midrule

      \multicolumn{10}{c}{\textit{Lightweight PLMs}} \\
  \midrule
\textsc{ALbert} &Top@1&79.64&79.33&79.59&78.96&83.86&80.95&79.88&79.23\\
&Te&1.9&1.9&1.9&2.0&52.6&38.0&80.5&76.2\\
\midrule

\textsc{DistilBert} &Top@1&81.57&81.11&80.97&81.52&85.41&82.59&81.74&\textbf{81.33}\\
&Te&\underline{0.7}&0.9&\underline{1.0}&\underline{0.8}&\underline{14.5}&\underline{19.4}&23.6&54.6\\
\midrule

\textsc{DistilRoberta}&Top@1&81.05&80.68&80.67&81.30&81.31&82.69&\textbf{82.06}&80.62\\
&Te&0.8&\underline{0.8}&\underline{1.0}&0.9&15.7&20.7&\underline{21.4}&\underline{28.2}\\
\midrule

      \multicolumn{10}{c}{\textit{Knowledge-enhanced PLMs}} \\
  \midrule
\textsc{Luke}&Top@1&\textbf{81.77}&\textbf{82.12}&\textbf{82.22}&\textbf{81.74}&85.50&\textbf{83.00}&82.00&80.60\\
&Te&1.2&1.5&1.5&1.7&27.7&34.9&37.0&47.9\\
\midrule

\textsc{Kepler}&Top@1&81.12&81.09&80.96&80.75&\textbf{85.60}&\textbf{83.00}&81.70&80.30\\
&Te&1.8&1.8&1.8&2.0&22.4&30.7&32.6&45.6\\
  \bottomrule
 \end{tabular}
 }
  \end{table}

As shown in Table \ref{RDResults}, there are significant differences in performance on RD between the two KGQA frameworks, which leads to differences in the final accuracy and efficiency of the two frameworks. All rows in Table \ref{RDResults} show that the \textsc{KGQAcl} is more efficient than the \textsc{KGQArr} because \textsc{KGQArr} needs to encode all candidate relations to and questions to calculate similarity, whereas \textsc{KGQAcl} only needs to encode questions. This is also why the increase in KG size significantly affects the accuracy and efficiency of \textsc{KGQArr}, yet it does not affect \textsc{KGQAcl}. Although the accuracy of \textsc{KGQArr} is significantly higher than that of \textsc{KGQAcl} on small-scale KG benchmark \textsc{SQs}, the former is less scalable than the latter. In addition, the knowledge-enhanced PLMs \textsc{Luke} and \textsc{Kepler} show the highest accuracy performance, which indicates the effectiveness of the knowledge-enhanced approach.

In general, both ED and RD modules significantly impact the final accuracy. ED and RD based on the \textsc{KGQArr} framework have a primary effect on the final efficiency, and they are most susceptible to changes in KG size. \textsc{KGQArr} has worse scalability than \textsc{KGQAcl} due to their different approaches to solving RD.

\begin{figure}
\begin{center}{
\includegraphics[width=0.95\textwidth]{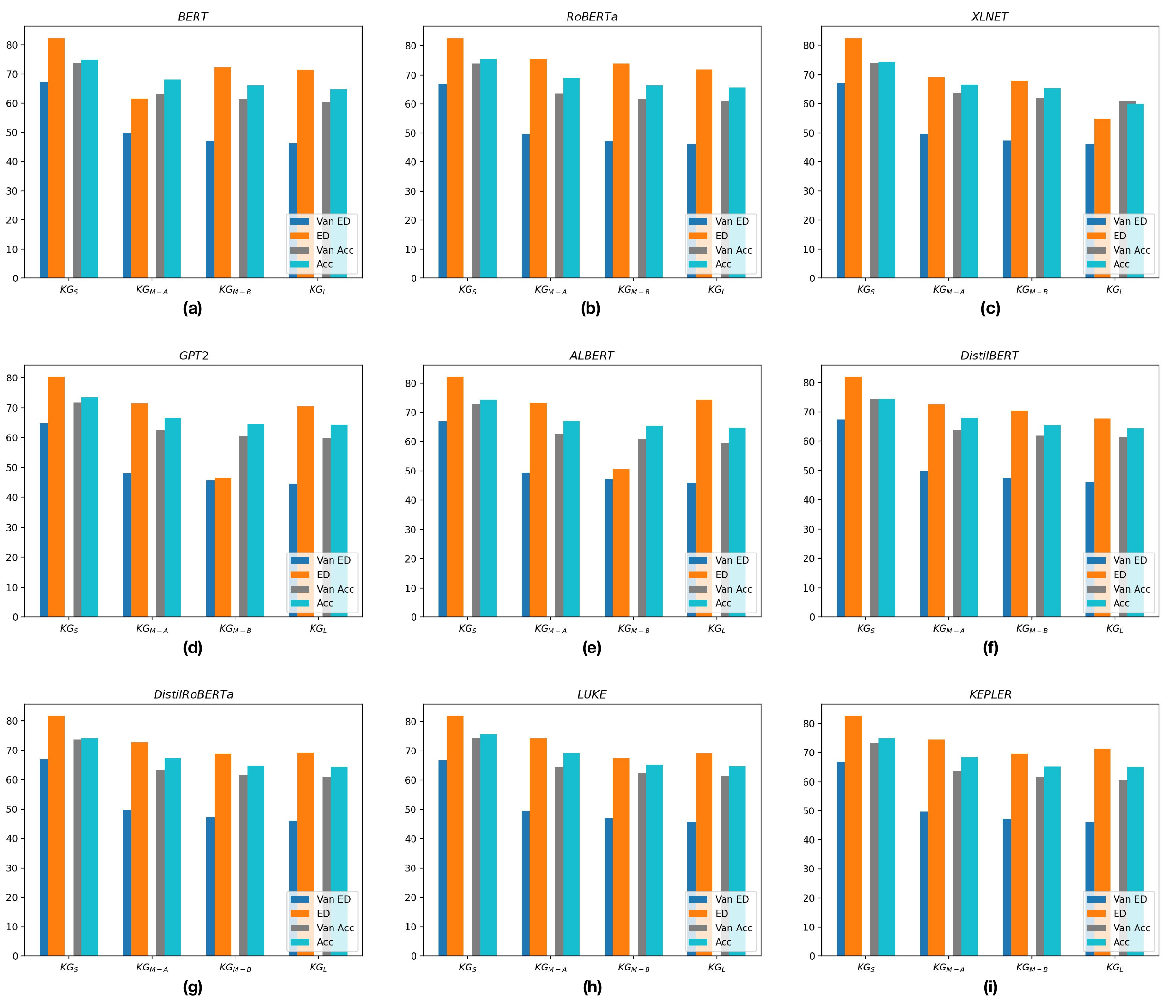}
\caption{Comparison of Top-1 recall of entity disambiguation and final accuracy results for various PLMs (a-i) based methods and vanilla method. Van ED and Van Acc denote entity disambiguation and the whole KGQA system using vanilla method, and ED denotes entity disambiguation and the whole KGQA system using PLM.} \label{VanilaEDandAcc}
}
\end{center}
\end{figure}

\begin{figure}
\begin{center}{
\includegraphics[width=0.8\textwidth]{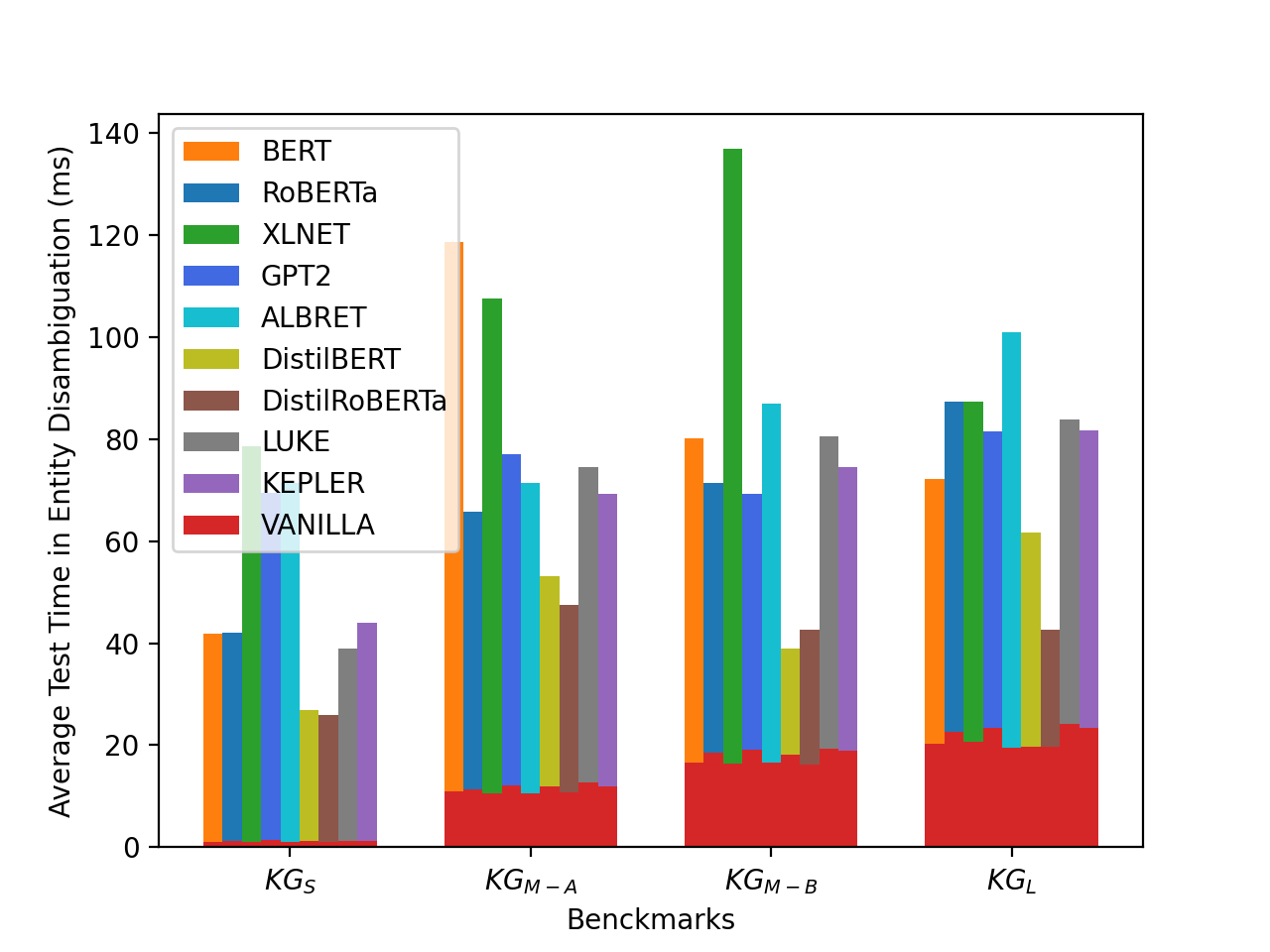}
\caption{The average test time of the various PLMs-based methods compared to the vanilla method used to resolve entity disambiguation in all benchmarks.} \label{VanTime}
}
\end{center}
\end{figure}

\subsubsection{Entity Disambiguation Using the Vanilla Method}

The analysis in Section \ref{sec4.3.1} demonstrates that PLMs-based entity disambiguation takes up the most time in the whole KGQA system. Given the high computational complexity of PLMs, we attempt to solve entity disambiguation using a vanilla method without any neural networks. We use only a simple linguistic approach, fuzzy matching, to rank all candidate entities (mentioned in Section \ref{sec3.2.2}). Specifically, we rank all candidate entities according to the Levenshtein Distance score between the entity name and the subject mention.

Table \ref{VanilaEDandAcc} shows the comparison results regarding the performance of vanilla methods and PLMs on entity disambiguation on all benchmarks, and their impact on the final accuracy. Compared to the vanilla method, all PLMs significantly improved the performance of entity disambiguation on all benchmarks (Van ED vs. ED). However, the improvements in the final accuracy of PLMs are not as significant in most cases (Van Acc vs. Acc). This is because the answer query module performs a weighted combination of candidate entities with scores and candidate relations with scores, which also screens out ambiguous entities to some extent. It is worth noting that \textsc{XLnet} improves entity disambiguation on the large-scale KG benchmark \textsc{KGl} (Fig. \ref{VanilaEDandAcc}(c)), yet is worse than the vanilla method in terms of final accuracy. This is because the large-scale KG contains too many noisy relations, leading the \textsc{XLnet}-based entity disambiguation model to assign a lower score to the ambiguous entities that some answer query modules can filter out. In addition, Fig.\ref{VanTime} shows the efficiency of the vanilla method compared to that of PLMs on entity disambiguation, the vanilla method takes much less time than PLMs. Therefore, PLMs-based entity disambiguation is time costly and has limited improvement in the final accuracy of KGQA. More importantly, using the \textsc{XLnet}-based entity disambiguation model can even reduce the final accuracy on large-scale KG benchmark \textsc{KGl}.

\begin{table}
 \centering
 \renewcommand\tabcolsep{1pt}
 \renewcommand{\arraystretch}{1.2}
 \caption{Overall accuracy and efficiency of different PLMs-based KGQA systems on WebQuestionSP (WB) and FreebaseQA (FBQ). Acc denotes accuracy (\%). Te refers to the average test time (ms). Bolded numbers indicate the highest accuracy. Underlined ones indicate the shortest time spent.}\label{WBFBQ}
 \scalebox{0.95}{
 \begin{tabular}{lc|cc|cc}
  \toprule
   \multicolumn{2}{c|}{\textbf{Frameworks}}&  \multicolumn{2}{|c|}{\textsc{KGQAcl}} & \multicolumn{2}{|c}{\textsc{KGQArr}}\\
  \midrule
   \multicolumn{2}{c|} {\diagbox{\textbf{PLMs}}{\textbf{Benckmarks}} }& \textsc{WQ} & \textsc{FBQ}&  \textsc{WQ} &  \textsc{FBQ} \\
  \midrule
      \multicolumn{6}{c}{\textit{Common PLMs}} \\
  \midrule
\textsc{Bert} (\textit{MLM)}&Acc &61.76&40.12&64.11&41.48\\
&Te&44.5&44.9&72.2&72.4\\
\midrule

\textsc{Roberta} (\textit{MLM)}&Acc&\textbf{62.32}&40.40&\textbf{64.59}&41.97\\
&Te&44.4&44.9&72.2&72.3\\
\midrule

\textsc{XLnet} (\textit{PeLM)}&Acc&61.46&39.80&63.87&41.15\\
&Te&85.1&85.1&119.8&117.7\\
\midrule

\textsc{Gpt2} (\textit{LM)}&Acc &60.49&39.03&18.11&5.09\\
&Te&48.3&48.5&76.3&76.5\\
\midrule

      \multicolumn{6}{c}{\textit{Lightweight PLMs}} \\
  \midrule
\textsc{ALbert} &Acc&61.47&39.83&63.89&41.16\\
&Te&55.1&55.3&104.9&104.9\\
\midrule

\textsc{DistilBert} &Acc&61.49&39.84&63.59&40.88\\
&Te&\underline{26.7}&\underline{26.9}&48.4&48.5\\
\midrule

\textsc{DistilRoberta}&Acc&61.05&39.43&62.57&39.36\\
&Te&28.9&28.0&\underline{43.2}&\underline{43.0}\\
\midrule

      \multicolumn{6}{c}{\textit{Knowledge-enhanced PLMs}} \\
  \midrule
\textsc{Luke}&Acc&62.31&\textbf{40.62}&64.52&\textbf{42.08}\\
&Te&41.2&41.3&67.5&67.7\\
\midrule

\textsc{Kepler}&Acc&62.02&40.29&64.46&42.02\\
&Te&48.8&48.8&68.8&69.0\\
  \bottomrule
 \end{tabular}
 }
  \end{table}

\subsection{Validation Beyond the SimpleQuestions Benchmarks}\label{Sec6.5}

In addition to the four benchmarks of the SimpleQuestions family (Sec.\ref{ConstructeBenchmarks}), we evaluated the accuracy and efficiency of all systems on the WebQuestionSP \cite{WebQSP} and FreebaseQA \cite{FreebaseQA} datasets.  
Both datasets adopt the large-scale KG, Freebase, as the resource and include a high proportion of simple questions (71.3\% in WebQuestionSP and 66.4\% FreebaseQA).
As these two datasets include numerous questions with multi-hop paths or multiple constraints, such as ``\textit{What character did Natalie Portman play in Star Wars?}'', we followed \cite{AnEmpirical} to pre-process these two datasets\footnote{The pre-processed datasets are available at https://github.com/aistairc/simple-qa-analysis.}. Specifically, we kept only simple questions that can be answered by a triple and questions with entities or predicates within FB2M.

Table \ref{WBFBQ} demonstrates the overall accuracy and efficiency results for all systems on WebQestionSP and FreebaseQA. Based on these results, we get similar conclusions to the experiments on SimpleQuestions. 
\textsc{Roberta} and \textsc{Luke} have the best performance. \textsc{Gpt2} perform the worst in terms of performance, especially in the \textsc{KGQArr} framework. Almost all PLMs based on the \textsc{KGQArr} framework have higher accuracy performance than those based on the \textsc{KGQAcl} framework but are more time-consuming. The two distillation-based PLMs, \textsc{DistilBert} and \textsc{DistilRoberta}, are far more efficient than the other PLMs. Furthermore, all systems performed poorly on FreebaseQA, with even the best, \textsc{Luke}, only achieving 42.08\% accuracy. After performing an error analysis, we found that FreebaseQA contained many mislabelled and unanswerable questions.

\subsection{ChatGPT for Zero-Shot KGQA}\label{Sec6.6}

\begin{table}
 \centering
 \renewcommand\tabcolsep{1pt}
 \renewcommand{\arraystretch}{1.2}
 \caption{Accuracy (\%) of ChatGPT, \textsc{KGQAcl} framework with GPT2 and \textsc{KGQArr} frameworks with the other PLMs on SimpleQuestions (SQ), WebQuestionSP (WB) and FreebaseQA (FBQ). Bolded numbers indicate the highest accuracy.}\label{CHATGPT}
 \scalebox{0.95}{
 \begin{tabular}{l|ccc}
\specialrule{0.8pt}{0pt}{1.2pt}
  \textbf{PLMs} & \multicolumn{3}{c}{\textbf{Benckmarks}} \\ 
  & \textbf{SQ} & \textbf{WQ}&  \textbf{FBQ}  \\
\specialrule{0.6pt}{1.2pt}{2.0pt}
\textsc{Bert}  &76.7&68.7& 41.7\\
\midrule

\textsc{Roberta} &\textbf{77.7}&69.0&42.0\\
\midrule

\textsc{XLnet} &75.7&68.0&41.0\\
\midrule

\textsc{Gpt2}  &72.7&64.3&39.0\\

\midrule
\textsc{ALbert} &76.3&66.7&41.3\\
\midrule

\textsc{DistilBert} &76.7&67.7&41.3\\
\midrule

\textsc{DistilRoberta}&76.0&67.0&40.3\\
\midrule
\textsc{Luke} &78.0&68.3&42.0\\
\midrule

\textsc{Kepler} &77.3&68.7&41.7\\

\specialrule{0.5pt}{1.8pt}{0.6pt}
\specialrule{0.5pt}{0.6pt}{1.5pt}

\textsc{ChatGPT} &29.3&\textbf{76.3}&\textbf{93.7}\\
  \bottomrule
 \end{tabular}
 }
  \end{table}

\begin{table}
 \centering
 \renewcommand\tabcolsep{1pt}
 \renewcommand{\arraystretch}{1.2}
 \caption{Six error types of ChatGPT on SimpleQuestions.}\label{CHATGPTError}
 \scalebox{1}{
 \begin{tabular}{p{11.4cm}}
\specialrule{1.0pt}{0pt}{1.2pt}
\textbf{1}. \textsc{Enumeration} (38.2\%) \\

\specialrule{0.6pt}{1.2pt}{1.2pt}
Question: \textit{what album is from the dance music genre?}  \\
ChatGPT: \textit{There are many albums that belong to the dance music genre, as it is a broad and diverse genre. Some popular dance music albums include:  * \textbf{Daft Punk - Homework (1997)} * \textbf{The Chemical Brothers - Dig Your Own Hole (1997)}...} \\
Gold: \textit{\textcolor{green}{love's crazy}} \\

\specialrule{1.0pt}{1.2pt}{1.2pt}

\textbf{2}. \textsc{ Wrong Answers} (26.9\%) \\
\specialrule{0.6pt}{1.2pt}{2.0pt}
Question: \textit{what is ivan dzerzhinsky profession ?}  \\
ChatGPT: \textit{Ivan Dzerzhinsky was a Soviet \textbf{statesman} and Bolshevik \textbf{revolutionary}...  } \\
Gold: \textit{\textcolor{green}{composer}} \\

\specialrule{1.0pt}{1.2pt}{1.2pt}
\textbf{3}. \textsc{Lack of Knowledge} (22.6\%) \\
\specialrule{0.6pt}{1.2pt}{2.0pt}
Question: \textit{which country does elizabeth larner have nationality in?}  \\
ChatGPT: \textit{I'm sorry, but I don't have any information on an individual named Elizabeth Larner and their nationality. Can you provide more context or details?} \\
Gold: \textit{\textcolor{green}{kingdom of great britain and northern ireland}} \\

\specialrule{1.0pt}{1.2pt}{1.2pt}
\textbf{4}. \textsc{Ambiguous Entities} (10.0\%) \\
\specialrule{0.6pt}{1.2pt}{2.0pt}
Question: \textit{who is the lyricist of forever?}  \\
ChatGPT: \textit{Forever is a popular song and there are several songs with the title "Forever." Can you provide more context or specify which "Forever" you're referring to? This would help me better answer your question.} \\
Gold: \textit{\textcolor{green}{ferr}} \\

\specialrule{1.0pt}{1.2pt}{1.2pt}
\textbf{5}. \textsc{Misunderstanding} (1.4\%) \\
\specialrule{0.6pt}{1.2pt}{2.0pt}
Question: \textit{who plays defender in soccer?}  \\
ChatGPT: \textit{In soccer, players who play in the defensive positions are referred to as defenders. The main responsibility of a defender is to prevent the opposing team from scoring goals by blocking shots, tackles, and interceptions...} \\
Gold: \textit{\textcolor{green}{bogdan bucurică}} \\

\specialrule{1.0pt}{1.2pt}{1.2pt}
\textbf{6}. \textsc{Dataset Problem} (0.9\%) \\
\specialrule{0.6pt}{1.2pt}{2.0pt}
Question: \textit{what is book was published by ?}  \\
ChatGPT: \textit{I'm sorry, but you haven't specified which book you are referring to. Can you please provide the name of the book so I can assist you better?} \\
Gold: \textit{\textcolor{green}{thirteenth tribe}} \\

  \bottomrule
 \end{tabular}
 }
  \end{table}
  
We conducted experiments to compare the performance between ChatGPT\footnote{The version of ChatGPT is Jau 30 Version, and the user's access times are limited. We have released the script for accessing ChatGPT.} and other PLMs on 300 sampling questions from SimpleQuestions, WebQuestionSP and FreebaseQA. Note that ChatGPT was under the zero-shot KGQA setting, while other PLMs were fine-tuned using the training set in a better-performing framework (\textsc{KGQAcl} framework with GPT2 and \textsc{KGQArr} frameworks with the other PLMs). The input of ChatGPT consists of the instruction (\textit{``Please answer the given question based on the context. The answers should be factual answers.''}) and the question, inspired by \cite{IsChatGPT}. After reading the entire input, the model generates the answer in the form of a piece of text\footnote{We have also tried to generate concise answers like entity names by specific instructions, but it leads to worse performance.}. For each question, the answers generated by ChatGPT were evaluated and cross-validated by two professionals with reference to gold answers.

Table \ref{CHATGPT} demonstrates that ChatGPT outperforms other PLMs by up to 12\% on WebQuestionSP and far surpasses other PLMs by up to 54\% on FreebaseQA. However, ChatGPT performs miserably on SimpleQuestions with an accuracy of only 29.3\%. We speculate that the discrepancy is caused by the different construction methods of these datasets. WebQuestionSP was derived from Google Suggest API, while FreebaseQA was scraped from trivia and quiz-league websites, which are still accessible. In contrast, SimpleQuestions were constructed by humans based on Freebase triples. Therefore, it is possible that ChatGPT has seen these questions or related texts due to its extremely large training corpus.

We further categorized the error cases of ChatGPT on SimpleQuestions as shown in Table \ref{CHATGPTError}. We consider the \textsc{Enumeration} type questions, which account for 38.2\%, as a category of errors due to the difficulty of verifying that all enumeration items are correct. Note that when enumeration items include the gold answers, we consider the answer as correct. That is, In type 1 \textsc{Enumeration}, ChatGPT's answer does not contain a golden answer. Even though we regard all the questions of Type 1 as correct, the accuracy of ChatGPT on SimpleQuestions is 56.3\%, which is still significantly inferior to other PLMs. \textsc{Wrong Answers} (account for 26.9\%) indicate that ChatGPT answers differ from gold answers. Besides, we noticed that ChatGPT even generates incorrect facts, also known as the hallucination problem \cite{AMultitask}. For example, in the second example in Table \ref{CHATGPTError}, this politician and revolutionary is actually \textit{Felix Dzerzhinsky} rather than \textit{Ivan Dzerzhinsky}. In addition, 22.6\% of the error cases are due to a lack of knowledge about the subject entity (i.e. \textsc{Lack of Knowledge}), and 10.0\% of the error cases are due to a lack of additional information to disambiguate the subject entity (i.e. \textsc{Ambiguous Entities}). 1.4\% of the errors are due to ChatGPT misunderstanding the semantics of the question (i.e. \textsc{Misunderstanding}). The poor quality of the question itself causes 0.9\% of the errors (i.e. \textsc{Dataset Problem}). These cases demonstrate that ChatGPT may generate factual errors and still lacks extensive factual knowledge since many subject entities cannot be identified.

\section{Conclusion and Future Works}\label{Sec6}

Due to the improved performance of PLMs on most NLP tasks, it has become a consensus to use PLMs as a skeleton to solve NLP tasks. In this paper, we investigate the application of PLMs to solve a knowledge-intensive task, namely knowledge graph question answering. We conduct comprehensive experiments to explore the accuracy and efficiency performance of PLMs on KGQA, as well as the scalability of PLMs as KG size increases. In addition, we compare the performance between ChatGPT and other PLMs on three KGQA datasets. We present a detailed analysis of these experimental results and draw some important conclusions regarding the use of PLMs in KGQA.

\begin{enumerate}[1.]

\item
\textsc{Roberta} and the knowledge-enhanced PLMs \textsc{Luke} and \textsc{Kepler} achieve the highest accuracy performance in the KGQA task. \textsc{Luke} and \textsc{Kepler} performed better on the small-scale KG benchmarks, and \textsc{Roberta} performed better on the large-scale KG benchmarks.

\item
Lightweight PLMs \textsc{DistilBert} and \textsc{DistilRoberta} with knowledge distillation technology significantly improve efficiency and have lower accuracy than other PLMs on the small-scale KG benchmarks. However, \textsc{DistilBert} and \textsc{DistilRoberta} exhibit the best scalability. As KG size increases, the gap between their accuracy and that of other PLMs is gradually eliminated.

\item
The accuracy of \textsc{XLnet} with permuted language modelling and \textsc{Gpt2} with language modelling is worse than that of PLMs with masked language modelling, especially the \textsc{KGQArr} framework based on \textsc{Gpt2}.

\item
The combined overall accuracy and efficiency results of KGQA show that PLMs-based entity disambiguation has no advantage over fuzzy matching-based entity disambiguation. 
Although the former is significantly better than the latter in the performance of entity disambiguation, the gap in accuracy between the two KGQA systems based on them is insignificant because the answer query module has the ability to disambiguate.

\item
ChatGPT shows superior performance on zero-shot WebQuestions and FreebaseQA, even significantly outperforming other PLMs with fine-tuning. We speculate that this is due to the fact that ChatGPT has seen a similar corpus during training, as it performs extremely poorly on manually constructed SimpleQuestions. The error case analysis suggests that ChatGPT may generate answers with incorrect facts and still lack knowledge since many subject entities cannot be identified.

\end{enumerate}

Further, we examine the overall results of the various PLMs on the subtasks of KGQA and obtain similar conclusions.
\textsc{Roberta} and \textsc{Bert} exhibit the best performance on the entity detection and entity disambiguation tasks, while the knowledge-enhanced PLMs \textsc{Luke} and \textsc{Kepler} show strong capabilities on the relation detection task.
\textsc{DistilBert} and \textsc{DistilRoberta} have a clear efficiency advantage and perform well on all tasks except for the entity disambiguation task, which is slightly inferior.
In addition, we find that the \textsc{KGQArr}-based systems are significantly less efficient than the \textsc{KGQAcl}-based systems. Furthermore, the higher accuracy of \textsc{KGQArr}-based systems than \textsc{KGQAcl}-based systems is only when the KG scale is small. As the KG scale increases, the former is gradually inferior to the latter, which indicates the poor scalability of the \textsc{KGQArr} framework.

In future work, we will extend the proposed simple KGQA framework to the multi-hop complex KGQA framework. We will also keep investigating the application of knowledge distillation and knowledge-enhanced PLMs in KGQA, as our experiments show them to be promising. In addition, we will follow up with artificial general intelligence models like ChatGPT and test them more carefully, especially in terms of efficiency.


\section*{Declarations}

\bmhead{Ethical Approval} Not applicable.

\bmhead{Competing interests} The authors declare that there are no competing interest regarding the publication of this article.

\bmhead{Authors' contributions} Nan Hu: Conceptualization, Methodology, Software, Writing - Original Draft. Yike Wu: Investigation, Software, Validation. Guilin Qi: Conceptualization, Methodology, Writing - review \& editing. Dehai Min: Software. Jiaoyan Chen: Writing - review \& editing. Jeff Z. Pan: Writing - review \& editing. Zafar Ali: Validation.

\bmhead{Funding} This work is supported by National Nature Science Foundation of China (No. U21A20488).

\bmhead{Availability of data and materials} All datasets and codes in this paper can be accessed from https://github.com/aannonymouuss/PLMs-in-Practical-KBQA.


\begin{thebibliography}{60}

\bibitem{ref_article1}
Manning, C.D.: Human Language Understanding \& Reasoning. In: Daedalus, pp. 127-138 (2022). https://doi.org/10.1162/daed\_a\_01905

\bibitem{ref_article2}
Mohammed, S., Shi, P., Lin, J.J.: Strong Baselines for Simple Question Answering over Knowledge Graphs with and without Neural Networks. In: Proceedings of the 2018 Conference of the North American Chapter of the Association for Computational Linguistics, pp. 291--296 (2018). https://doi.org/10.18653/v1/n18-2047

\bibitem{ref_article3}
Lukovnikov, D., Fischer, A., Lehmann, J.: Pretrained Transformers for Simple Question Answering over Knowledge Graphs. In: 18th International Semantic Web Conference, pp. 470-486 (2019). https://doi.org/10.1007/978-3-030-30793-6\_27

\bibitem{ref_article4}
Golub, D., He, X.: Character-Level Question Answering with Attention. In: Proceedings of the 2016 Conference on Empirical Methods in Natural Language Processing, pp. 1598-1607 (2016). https://doi.org/10.18653/v1/d16-1166

\bibitem{ref_article5}
Petrochuk, M., Zettlemoyer, L.: SimpleQuestions Nearly Solved: A New Upperbound and Baseline Approach. In: Proceedings of the 2018 Conference on Empirical Methods in Natural Language Processing, pp. 554-558 (2018). https://doi.org/10.18653/v1/d18-1051

\bibitem{ref_article6}
Türe, F., Jojic, O.: No Need to Pay Attention: Simple Recurrent Neural Networks Work! In: Proceedings of the 2017 Conference on Empirical Methods in Natural Language Processing, pp. 2866-2872 (2017). https://doi.org/10.18653/v1/d17-1307

\bibitem{Yu2017}
Yu, M., Yin, W., Hasan, K.S., Santos, C.N., Xiang, B., Zhou, B.: Improved Neural Relation Detection for Knowledge Base Question Answering. In: Proceedings of the 55th Annual Meeting of the Association for Computational Linguistics, pp. 571-581 (2017). https://doi.org/10.18653/v1/P17-1053

\bibitem{ref_article8}
Cui, H., Peng, T., Feng, L., Bao, T., Liu, L.: Simple Question Answering over Knowledge Graph Enhanced by Question Pattern Classification. In: Knowl. Inf. Syst., pp. 2741-2761  (2021). https://doi.org/10.1007/s10115-021-01609-w

\bibitem{ref_article9}
Lukovnikov, D., Fischer, A., Lehmann, J., Auer, S.: Neural Network-based Question Answering over Knowledge Graphs on Word and Character Level. In: Proceedings of the 26th International Conference on World Wide Web, pp. 1211-1220 (2017). https://doi.org/10.1145/3038912.3052675

\bibitem{ref_article10}
Hao, Y., Liu, H., He, S., Liu, K., Zhao, J.: Pattern-revising Enhanced Simple Question Answering over Knowledge Bases. In: Proceedings of the 27th International Conference on Computational Linguistics, pp. 3272-3282 (2018). https://aclanthology.org/C18-1277/

\bibitem{ref_article11}
Yin, W., Yu, M., Xiang, B., Zhou, B., Schütze, H.: Simple Question Answering by Attentive Convolutional Neural Network. In: Proceedings of the 26th International Conference on Computational Linguistics, pp. 1746-1756 (2016). https://aclanthology.org/C16-1164/

\bibitem{ref_article12}
Zhao W., Chung T., Goyal AK., Metallinou A.: Simple question answering with subgraph ranking
and joint-scoring. In: Proceedings of the 2019 Conference of the North American Chapter of the Association for Computational Linguistics, pp. 324–334 (2019). https://doi.org/10.18653/v1/n19-1029

\bibitem{ref_article13}
Hao, Y., Zhang, Y., Liu, K., He, S., Liu, Z., Wu, H., Zhao, J.: An End-to-End Model for Question Answering over Knowledge Base with Cross-Attention Combining Global Knowledge. In: Proceedings of the 55th Annual Meeting of the Association for Computational Linguistics, pp. 221–231 (2017). https://doi.org/10.18653/v1/P17-1021

\bibitem{ref_article14}
Luo, D., Su, J., Yu, S.: A BERT-based Approach with Relation-aware Attention for Knowledge Base Question Answering. In: 2020 International Joint Conference on Neural Networks, pp. 1-8 (2020). https://doi.org/10.1109/IJCNN48605.2020.9207186

\bibitem{ref_article15}
Dai, Z., Li, L., Xu, W.: CFO: Conditional Focused Neural Question Answering with Large-scale Knowledge Bases. In: Proceedings of the 54th Annual Meeting of the Association for Computational Linguistics, pp. 800–810 (2016). https://doi.org/10.18653/v1/p16-1076

\bibitem{ref_article16}
Lan, Y., Wang, S., Jiang, J.: Knowledge Base Question Answering With a Matching-Aggregation Model and Question-Specific Contextual Relations. In: IEEE ACM Trans. Audio Speech Lang. Process., pp. 1629-1638 (2019). https://doi.org/10.1109/TASLP.2019.2926125

\bibitem{Transformer}
Vaswani, A., Shazeer, N.M., Parmar, N., Uszkoreit, J., Jones, L., Gomez, A.N., Kaiser, L., Polosukhin, I.: Attention is All you Need. In: Advances in Neural Information Processing Systems, pp. 5998-6008 (2017)

\bibitem{LSTM}
Hochreiter, S., Schmidhuber, J.: Long Short-Term Memory. In: Neural Computation, pp. 1735-1780 (1997)

\bibitem{CNN}
Krizhevsky, A., Sutskever, I., Hinton, G.E.: ImageNet classification with deep convolutional neural networks. In: Communications of the ACM, pp. 84 - 90 (2017). http://doi.org/10.1145/3065386

\bibitem{BERT}
Devlin, J., Chang, M., Lee, K., Toutanova, K.: BERT: Pre-training of Deep Bidirectional Transformers for Language Understanding. In: Proceedings of the 2019 Conference of the North American Chapter of the Association for Computational Linguistics, pp. 4171-4186 (2019). https://doi.org/10.18653/v1/n19-1423

\bibitem{RoBERTa}
Liu, Y., Ott, M., Goyal, N., Du, J., Joshi, M., Chen, D., Levy, O., Lewis, M., Zettlemoyer, L., Stoyanov, V.: RoBERTa: A Robustly Optimized BERT Pretraining Approach. In: ArXiv, 1907.11692 (2019)

\bibitem{ALBERT2021}
Lan, Z., Chen, M., Goodman, S., Gimpel, K., Sharma, P., Soricut, R.: ALBERT: A Lite BERT for Self-supervised Learning of Language Representations. In: 8th International Conference on Learning Representations (2020)

\bibitem{KGEmbedding2019}
Huang, X., Zhang, J., Li, D., Li, P.: Knowledge Graph Embedding Based Question Answering. In: Proceedings of the Twelfth ACM International Conference on Web Search and Data Mining, pp. 105-113 (2019). https://doi.org/10.1145/3289600.3290956

\bibitem{ref_article25}
Bordes, A., Usunier, N., Chopra, S., Weston, J.: Large-scale Simple Question Answering with Memory Networks. In: ArXiv, 1506.02075 (2015)

\bibitem{Word2Vec}
Mikolov, T., Sutskever, I., Chen, K., Corrado, G.S., Dean, J.: Distributed Representations of Words and Phrases and their Compositionality. In: Advances in Neural Information Processing Systems, pp. 3111-3119 (2013)

\bibitem{GloVe}
Pennington, J., Socher, R., Manning, C.D.: GloVe: Global Vectors for Word Representation. In: Proceedings of the 2014 Conference on Empirical Methods in Natural Language Processing, pp. 1532-1543 (2014). https://doi.org/10.3115/v1/d14-1162

\bibitem{ref_article28}
Reimers N. , Gurevych I.: Sentence-BERT: Sentence Embeddings using Siamese BERT-Networks. In: Proceedings of the 2019 Conference on Empirical Methods in Natural Language Processing, pp. 3980-3990 (2019). https://doi.org/10.18653/v1/D19-1410

\bibitem{Li2020}
Li, B.Z., Min, S., Iyer, S., Mehdad, Y., Yih, W.: Efficient One-Pass End-to-End Entity Linking for Questions. In: Proceedings of the 2020 Conference on Empirical Methods in Natural Language Processing, pp. 6433-6441 (2020). https://doi.org/10.18653/v1/2020.emnlp-main.522

\bibitem{Wu2020}
Wu, L.Y., Petroni, F., Josifoski, M., Riedel, S., Zettlemoyer, L.: Scalable Zero-shot Entity Linking with Dense Entity Retrieval. In: Proceedings of the 2020 Conference on Empirical Methods in Natural Language Processing, pp. 6397-6407 (2020). https://doi.org/10.18653/v1/2020.emnlp-main.519

\bibitem{Chen2020}
Chen, S., Wang, J., Jiang, F., Lin, C.: Improving Entity Linking by Modeling Latent Entity Type Information. In: The Thirty-Fourth AAAI Conference on Artificial Intelligence, pp. 7529-7537 (2020)

\bibitem{Oliya2021}
Oliya, A., Saffari, A., Sen, P., Ayoola, T.: End-to-End Entity Resolution and Question Answering Using Differentiable Knowledge Graphs. In: Proceedings of the 2021 Conference on Empirical Methods in Natural Language Processing, pp. 4193-4200 (2021). https://doi.org/10.18653/v1/2021.emnlp-main.345

\bibitem{Wang2021}
Wang, Z., Ng, P.K., Nallapati, R., Xiang, B.: Retrieval, Re-ranking and Multi-task Learning for Knowledge-Base Question Answering. In: Proceedings of the 16th Conference of the European Chapter of the Association for Computational Linguistics, pp. 347-357 (2021). https://doi.org/10.18653/v1/2021.eacl-main.26

\bibitem{Yamada2020}
Yamada, I., Asai, A., Shindo, H., Takeda, H., Matsumoto, Y.: LUKE: Deep Contextualized Entity Representations with Entity-aware Self-attention. In: Proceedings of the 2020 Conference on Empirical Methods in Natural Language Processing, pp. 6442-6454 (2020). https://doi.org/10.18653/v1/2020.emnlp-main.523

\bibitem{Zhang2022}
Zhang, T., Wang, C., Hu, N., Qiu, M., Tang, C., He, X., Huang, J.: DKPLM: Decomposable Knowledge-enhanced Pre-trained Language Model for Natural Language Understanding. In: Thirty-Sixth AAAI Conference on Artificial Intelligence, pp. 11703-11711 (2022)

\bibitem{Zhang2019}
Zhang, Z., Han, X., Liu, Z., Jiang, X., Sun, M., Liu, Q.: ERNIE: Enhanced Language Representation with Informative Entities. In: Proceedings of the 57th Conference of the Association for Computational Linguistics, pp. 1441-1451 (2019). https://doi.org/10.18653/v1/p19-1139

\bibitem{KEPLER2021}
Wang, X., Gao, T., Zhu, Z., Liu, Z., Li, J., Tang, J.: KEPLER: A Unified Model for Knowledge Embedding and Pre-trained Language Representation. In: Transactions of the Association for Computational Linguistics, 9, pp. 176-194. (2021). https://doi.org/10.1162/tacl\_a\_00360

\bibitem{Peters2019}
Peters, M.E., Neumann, M., RobertL.Logan, I., Schwartz, R., Joshi, V., Singh, S., Smith, N.A.: Knowledge Enhanced Contextual Word Representations. In: Proceedings of the 2019 Conference on Empirical Methods in Natural Language Processing, pp. 43-54 (2019). https://doi.org/10.18653/v1/D19-1005

\bibitem{Bollacker2008}
Bollacker, K.D., Evans, C., Paritosh, P.K., Sturge, T., Taylor, J.: Freebase: a collaboratively created graph database for structuring human knowledge. In: Proceedings of the ACM SIGMOD International Conference on Management of Data, pp. 1247-1250 (2008). https://doi.org/10.1145/1376616.1376746

\bibitem{Danny2020}
Danny Sullivan.: A reintroduction to our Knowledge Graph and knowledge panels. https://blog.google/products/search/about-knowledge-graph-and-knowledge-panels/ (2020). Accessed 3 October 2022

\bibitem{Lan2020}
Lan, Y., Jiang, J.: Query Graph Generation for Answering Multi-hop Complex Questions from Knowledge Bases. In: Proceedings of the 58th Annual Meeting of the Association for Computational Linguistics, pp. 969-974 (2020). https://doi.org/10.18653/v1/2020.acl-main.91

\bibitem{Gu2021}
Gu, Y., Kase, S.E., Vanni, M.T., Sadler, B.M., Liang, P., Yan, X., Su, Y.: Beyond I.I.D.: Three Levels of Generalization for Question Answering on Knowledge Bases. In: Proceedings of the Web Conference, pp. 3477-3488 (2021). https://doi.org/10.1145/3442381.3449992

\bibitem{Ye2022}
Ye, X., Yavuz, S., Hashimoto, K., Zhou, Y., Xiong, C.: RNG-KBQA: Generation Augmented Iterative Ranking for Knowledge Base Question Answering. In: Proceedings of the 60th Annual Meeting of the Association for Computational Linguistics, pp. 6032-6043 (2022). https://doi.org/10.18653/v1/2022.acl-long.417

\bibitem{Gu2022}
Gu, Y., Su, Y.: ArcaneQA: Dynamic Program Induction and Contextualized Encoding for Knowledge Base Question Answering. In: Proceedings of the 29th International Conference on Computational Linguistics, pp. 1718-1731 (2022)

\bibitem{Chen2021}
Chen, S., Liu, Q., Yu, Z., Lin, C., Lou, J., Jiang, F.: ReTraCk: A Flexible and Efficient Framework for Knowledge Base Question Answering. In: Proceedings of the 59th Annual Meeting of the Association for Computational Linguistics, pp. 325-336 (2021). https://doi.org/10.18653/v1/2021.acl-demo.39

\bibitem{Qin2021}
Qin, K., Li, C., Pavlu, V., Aslam, J.A.: Improving Query Graph Generation for Complex Question Answering over Knowledge Base. In: Proceedings of the 2021 Conference on Empirical Methods in Natural Language Processing, pp. 4201-4207 (2021). https://doi.org/10.18653/v1/2021.emnlp-main.346

\bibitem{unifiedSKG}
Xie, T., Wu, C., Shi, P., Zhong, R., Scholak, T., Yasunaga, M., Wu, C., Zhong, M., Yin, P., Wang, S.I., Zhong, V., Wang, B., Li, C., Boyle, C., Ni, A., Yao, Z., Radev, D., Xiong, C., Kong, L., Zhang, R., Smith, N.A., Zettlemoyer, L., Yu, T.: UnifiedSKG: Unifying and Multi-Tasking Structured Knowledge Grounding with Text-to-Text Language Models. In: ArXiv, 2201.05966 (2022)

\bibitem{TinyBERT}
Jiao, X., Yin, Y., Shang, L., Jiang, X., Chen, X., Li, L., Wang, F., Liu, Q.: TinyBERT: Distilling BERT for Natural Language Understanding. In: Findings of the Association for Computational Linguistics, pp. 4163-4174 (2020). https://doi.org/10.18653/v1/2020.findings-emnlp.372

\bibitem{DistilBERT}
Sanh, V., Debut, L., Chaumond, J., Wolf, T.: DistilBERT, a distilled version of BERT: smaller, faster, cheaper and lighter. In: ArXiv, 1910.01108 (2019)

\bibitem{GPT}
Radford, A., Narasimhan, K., Salimans, T., Sutskever, I.: Improving language understanding by generative pre-training. https://s3-us-west-2.amazonaws.com/openai-assets/research-covers/language-unsupervised/language\_understanding\_paper.pdf (2018). Accessed 4 October 2022

\bibitem{XLnet}
Yang, Z., Dai, Z., Yang, Y., Carbonell, J.G., Salakhutdinov, R., Le, Q.V.: XLNet: Generalized Autoregressive Pretraining for Language Understanding. In: Advances in Neural Information Processing Systems, pp. 5754-5764 (2019)

\bibitem{AReview}
Zhang, C., Lai, Y., Feng, Y., Zhao, D.: A review of deep learning in question answering over knowledge bases. In: AI Open, pp. 205-215 (2021)

\bibitem{ComplexKnowledge}
Lan, Y., He, G., Jiang, J., Jiang, J., Zhao, W.X., Wen, J.: Complex Knowledge Base Question Answering: A Survey. In: IEEE TKDE (2021)

\bibitem{KnowledgeBase}
Gu, Y., Pahuja, V., Cheng, G., Su, Y.: Knowledge Base Question Answering: A Semantic Parsing Perspective. ArXiv, In: ArXiv, 2209.04994 (2022)

\bibitem{Pre-trainedModels}
Qiu, X., Sun, T., Xu, Y., Shao, Y., Dai, N., Huang, X.: Pre-trained models for natural language processing: A survey. In: Science China Technological Sciences, pp. 1872 - 1897 (2020)

\bibitem{AnEmpirical}
Han, N., Topic, G., Noji, H., Takamura, H., Miyao, Y.: An empirical analysis of existing systems and datasets toward general simple question answering. In: COLING, pp. 5321-5334 (2020)

\bibitem{FreebaseQA}
Jiang, K., Wu, D., Jiang, H.: FreebaseQA: A New Factoid QA Data Set Matching Trivia-Style Question-Answer Pairs with Freebase. In: North American Chapter of the Association for Computational Linguistics, pp. 318-323 (2019)

\bibitem{WebQSP}
Yih, W., Richardson, M., Meek, C., Chang, M., Suh, J.: The Value of Semantic Parse Labeling for Knowledge Base Question Answering. In: Annual Meeting of the Association for Computational Linguistics (2016)

\bibitem{ImprovingCore}
Hu, N., Bi, S., Qi, G., Wang, M., Hua, Y., Shen, S.: Improving Core Path Reasoning for the Weakly Supervised Knowledge Base Question Answering. In: DASFAA, pp. 162-170 (2022)

\bibitem{SubgraphRetrieval}
Zhang, J., Zhang, X., Yu, J., Tang, J., Tang, J., Li, C., Chen, H.: Subgraph Retrieval Enhanced Model for Multi-hop Knowledge Base Question Answering. In: Annual Meeting of the Association for Computational Linguistics, pp. 5773-5784 (2022)

\bibitem{Case-based}
Das, R., Zaheer, M., Thai, D.N., Godbole, A., Perez, E., Lee, J., Tan, L., Polymenakos, L., McCallum, A.: Case-based Reasoning for Natural Language Queries over Knowledge Bases. In: Conference on Empirical Methods in Natural Language Processing, pp. 9594-9611 (2021)

\bibitem{RNG-KBQA}
Ye, X., Yavuz, S., Hashimoto, K., Zhou, Y., Xiong, C.: RNG-KBQA: Generation Augmented Iterative Ranking for Knowledge Base Question Answering. In: Annual Meeting of the Association for Computational Linguistics, pp. 6032-6043 (2021)

\bibitem{IsChatGPT}
Qin, C., Zhang, A., Zhang, Z., Chen, J., Yasunaga, M., Yang, D.: Is ChatGPT a General-Purpose Natural Language Processing Task Solver? In: ArXiv, 2302.06476 (2023)

\bibitem{AMultitask}
Bang, Y., Cahyawijaya, S., Lee, N., Dai, W., Su, D., Wilie, B., Lovenia, H., Ji, Z., Yu, T., Chung, W., Do, Q.V., Xu, Y., Fung, P.: A Multitask, Multilingual, Multimodal Evaluation of ChatGPT on Reasoning, Hallucination, and Interactivity. In: ArXiv, 2302.04023 (2023)

\bibitem{HowClose}
Guo, B., Zhang, X., Wang, Z., Jiang, M., Nie, J., Ding, Y., Yue, J., Wu, Y.: How Close is ChatGPT to Human Experts? Comparison Corpus, Evaluation, and Detection. In: ArXiv, 2301.07597 (2023)

\bibitem{DeepReinforcement}
Christiano, P.F., Leike, J., Brown, T.B., Martic, M., Legg, S., Amodei, D.: Deep Reinforcement Learning from Human Preferences. In: Neural Information Processing Systems, pp. 4299-4307 (2017)

\end{thebibliography}
\end{document}